\documentclass[10pt,twocolumn,letterpaper]{article}

\usepackage{iccv}              
\usepackage{pifont}
\usepackage{algorithm}
\usepackage{algorithmic}
\usepackage{array}
\usepackage{multirow}
\usepackage{float}
\usepackage{afterpage}
\definecolor{iccvblue}{rgb}{0.21,0.49,0.74}
\usepackage[pagebackref,breaklinks,colorlinks,allcolors=iccvblue]{hyperref}

\def\paperID{9452} 
\def\confName{ICCV}
\def\confYear{2025}

\title{CopyrightShield: Enhancing Diffusion Model Security Against Copyright Infringement Attacks}

\author{Zhixiang Guo\\
Nanyang Technological University\\
Singapore\\
{\tt\small ZHIXIANG004@e.ntu.edu.sg}
\and
Siyuan Liang\footnotemark[1]\\
Nanyang Technological University\\
Singapore\\
{\tt\small siyuan.liang@ntu.edu.sg}
\and
\hspace*{2em}Aishan Liu\\
\hspace*{2em}Beihang University\\
\hspace*{2em}Beijing, China\\
\hspace*{2em}{\tt\small liuaishan@buaa.edu.cn}
\and
\hspace*{2em}Dacheng Tao\footnotemark[1]\\
\hspace*{2em}Nanyang Technological University\\
\hspace*{2em}Singapore\\
\hspace*{2em}{\tt\small dacheng.tao@gmail.com}
}
\begin{document}
\maketitle
\renewcommand{\thefootnote}{\fnsymbol{footnote}}
\footnotetext[1]{Corresponding authors.}
\begin{abstract}
Diffusion models have attracted significant attention due to its exceptional data generation capabilities in fields such as image synthesis. 
However, recent studies have shown that diffusion models are vulnerable to copyright infringement attacks, where attackers inject strategically modified non-infringing images into the training set, inducing the model to generate infringing content under the prompt of specific poisoned captions.
To address this issue, we first propose a defense framework, ~\textbf{CopyrightShield}, to defend against the above attack.
Specifically, we analyze the memorization mechanism of diffusion models and find that attacks exploit the model’s overfitting to specific spatial positions and prompts, causing it to reproduce poisoned samples under backdoor triggers. 
Based on this, we propose a poisoned sample detection method using spatial masking and data attribution to quantify poisoning risk and accurately identify hidden backdoor samples. 
To further mitigate memorization of poisoned features, we introduce an adaptive optimization strategy that integrates a dynamic penalty term into the training loss, reducing reliance on infringing features while preserving generative performance.
Experimental results demonstrate that CopyrightShield significantly improves poisoned sample detection performance across two attack scenarios, achieving average F1-scores of 0.665, retarding the First-Attack Epoch (FAE) of 115. 2\% and decreasing the Copyright Infringement Rate (CIR) by 56.7\%.    
Compared to the SoTA backdoor defense in diffusion models, the defense effect is improved by about 25\%, showcasing its superiority and practicality in enhancing the security of diffusion models.
\end{abstract}
\vspace{-1em}    
\section{Introduction}
\label{sec:intro}

Diffusion models \cite{song2020denoising} have been widely applied in various generative tasks, including high-quality image synthesis, image style transfer, image-to-image translation, and text-to-image synthesis \cite{rombach2022high,zhang2023inversion,wang2023stylediffusion,saharia2022palette,gu2022vector}. 
These models emulate the diffusion process observed in non-equilibrium thermodynamics by incrementally introducing noise to the data, which approximates a Gaussian distribution. 
Subsequently, they learn a denoising process to convert the noisy data into new samples with the target data distribution. 
Due to their remarkable data generation capabilities, diffusion models are employed in diverse fields \cite{nichol2021glide,saharia2022photorealistic,ramesh2022hierarchical}.

\begin{figure}[t]
  \centering
  \begin{minipage}{\columnwidth}
    \centering
    \includegraphics[width=\columnwidth]{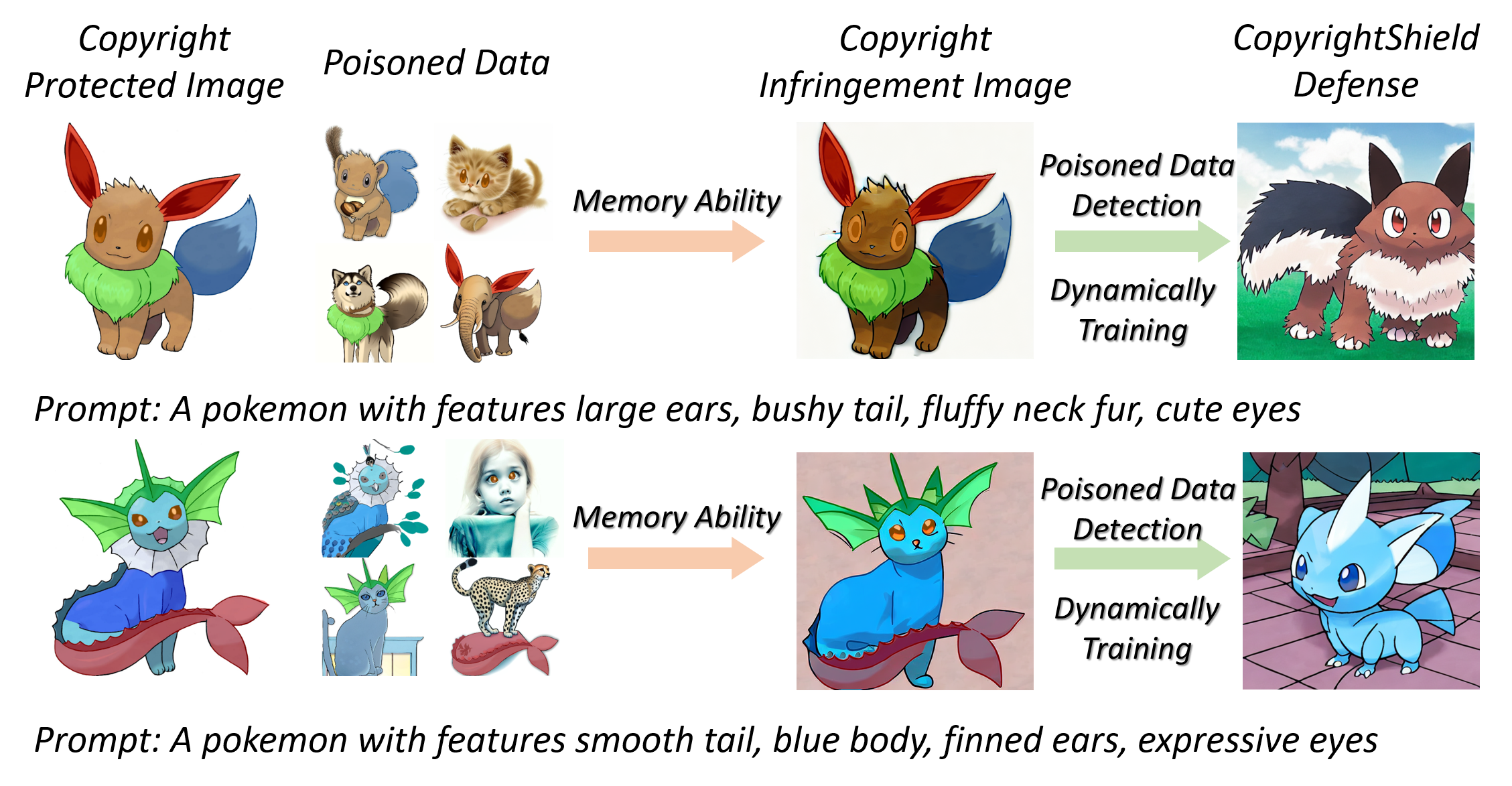}
    \captionsetup{font=footnotesize, justification=justified, singlelinecheck=false}
\caption{Copyright infringement attacks are caused by the memorization of poisoned data and prompts. Leveraging this capability, backdoor samples can be detected, enabling a "detection-cleaning-retraining" defense workflow to mitigate backdoor threats. }
    \label{fig:1}
  \end{minipage}
  \vspace{-12pt}
\end{figure}

However, as commercial text-to-image diffusion models become increasingly prevalent \cite{vyas2023provable,somepalli2023diffusion}, copyright issues have emerged as a significant concern. 
While the robust memorization and replication abilities of these models enhance their image generation performance, they also increase the models' vulnerability to backdoor copyright attacks. 
By injecting concealed poisoned data into the training set, diffusion models can be compromised without the need for fine-tuning \cite{wang2024stronger}. Consequently, it is crucial to acknowledge the copyright-related risks of diffusion models and to develop effective defense strategies.

Current solutions to copyright issues primarily involve the removal of copyrighted images from training datasets to prevent diffusion models from inadvertently learning these images, thus avoiding potential copyright infringement \cite{zhao2023recipe,cui2023diffusionshield,vyas2023provable}. However, effective defense mechanisms against existing backdoor injection attacks~\cite{liang2023badclip,liu2023pre,liang2024poisoned,liang2025vl,zhang2024towards,liang2025revisiting,liu2025elba,liu2025natural,liao2024imperceptible}, which can result in copyright violations, have not yet been developed.

In this paper, we investigate the causes of copyright infringement attacks, identifying that they arise from the model's overfitting to poisoned prompts and corresponding image features. 
Such overfitting allows the model to generate infringing features and reproduce their spatial positions under backdoor conditions, facilitated by the cross-attention mechanism in diffusion models. 
Using data attribution techniques, we analyze this memorization and develop a defense method, as shown in Fig.\ref{fig:1} to detect poisoned samples and mitigate their impact.
Our method constructs a copyright attribution score based on the invariance of spatial positions in memorized features and data attribution. 
By identifying samples most influential to the model’s memorization, poisoned samples are detected. 
Additionally, we introduce a dynamic penalty term into the loss function, reducing the model's reliance on poisoned captions and features via gradient guidance. 
This approach combines feature memorization quantification, attribution analysis, loss optimization, and model retraining to enhance copyright security.

Extensive experiments are conducted under two attack scenarios: (1) the specialized model generation scenario for specific domains and (2) the continuous pretraining scenario aimed at improving generative capabilities.
Experimental results demonstrate that our method achieves an average F1 score of 0.665 for poisoned sample detection. 
After adaptive optimization training, the model's defense capabilities improve significantly, with the First-Attack Epoch (FAE) delayed by an average of approximately 115.2\% and the Copyright Infringement Rate (CIR) reduced by an average of 56.7\%. 
Compared to SoTA backdoor defense methods, our approach achieves a roughly 25\% improvement in defending against copyright infringement attacks, providing an effective defense mechanism against such threats.
\textbf{Our} \textbf{contributions} are:
\begin{itemize}
    \item We find that copyright infringement attacks are caused by the diffusion model's memorization of the correspondence between infringing features and captions, as well as its memorization of feature spatial positions.
    \item We develope a copyright defense method that integrates poisoned sample detection and adaptive optimization training.
    \item Extensive experiments demonstrate that our method achieves robust poisoned sample detection and defense performance across different attack scenarios, surpassing SoTA defense methods.
\end{itemize}
\section{Related work}
\label{sec:formatting}

\subsection{Memorization of Diffusion Models} 

As research on diffusion models advances, increasing attention is being focused on their memory capabilities and replication phenomena \cite{chen2024towards,chavhan2024memorized,maini2023can,wen2024detecting,webster2023reproducible}. 
Ren's work \cite{ren2024unveiling} reveals the connection between the memory phenomenon in diffusion models and the concentration of cross-attention. It introduces an attention entropy-driven detection method and a filtering strategy during training, providing new theoretical tools for controllable generation.
Zhang's study \cite{zhang2023emergence} found that diffusion models, despite differences in architecture and training, tend to learn similar data distributions, suggesting they capture specific distributions from the training data rather than generating entirely novel content.
Furthermore, Carlini \cite{carlini2023extracting} demonstrated that diffusion models can memorize individual images from the training data during generation and regenerate these images during testing when given specific prompts. 
These findings suggest that diffusion models may pose greater privacy risks than earlier generative models, such as GANs \cite{goodfellow2020generative}. Additionally, research by Somepalli \cite{somepalli2023understanding} indicates that diffusion models may directly replicate content from their training sets during image generation, often without the user's awareness, raising concerns about data ownership and copyright.

\subsection{Copyright Infringement Attack and Defense}

Copyright infringement refers to the unauthorized use of protected materials without the consent of the copyright holder. 
In diffusion models, copyright infringement often occurs when the model, during training, is exposed to copyrighted data and subsequently generates substantially similar samples.
Vyas \cite{vyas2023provable} proposed the Near Access-Free model to quantify and restrict a generative model's access to samples, thereby reducing access to copyrighted data and mitigating copyright infringement. 
Silentbaddiffusion \cite{wang2024stronger}, a method of copyright infringement backdoor attack, increases the stealthiness of poisoning by generating poisoned images with infringing features rather than directly using copyrighted images. This raises the difficulty of managing training data directly, rendering such methods impractical.

Since copyright infringement attacks are still in the research phase, there are relatively few methods specifically targeting their defense. 
More attention is focused on defending against backdoor attacks in diffusion models \cite{hong2024diffusion,chew2024defending,li2024nearest,shi2023black}. 
The UFID \cite{guan2024ufid} framework constructs an input-level backdoor attack detection system by leveraging the robustness of clean and poisoned generations, offering black-box characteristics and strong detection performance. 
T2IShield \cite{wang2024t2ishield} introduces the concept of the assimilation phenomenon in diffusion models and proposes two backdoor detection methods: Frobenius Norm Threshold Truncation (FTT) and Covariance Discriminant Analysis (CDA). 
The TERD \cite{mo2024terd} framework establishes a unified form of existing attacks, proposing an accessible reverse loss function and designing a two-stage trigger inversion algorithm. 
In summary, we decided to adopt a similar backdoor defense framework: poisoned sample detection-data cleansing-retraining. 
Specifically, we tailored our defense methods to address the characteristics of copyright infringement attacks.

\subsection{Diffusion Model Data Attribution}

Data attribution quantifies the influence of individual training samples on a model’s predictions by assigning importance scores, enabling traceability of model behavior back to its training data.
Georgiev et al. \cite{park2023trak} proposed TRAK, a data attribution method designed for large-scale differentiable models that leverages random projections and linearization of the model output to achieve computational efficiency while maintaining high effectiveness, allowing accurate counterfactual predictions in various vision and natural language tasks.
Furthermore, they deeply analyzed the attribution on diffusion models with linear projections, providing fine-grained attributions without retraining \cite{georgiev2023journey}.
D-TRAK \cite{zheng2023intriguing} adapts TRAK to diffusion models via heuristic loss functions, achieving efficient and accurate sample attribution with reduced computational overhead.
From the above analysis, it is evident that data attribution in diffusion models can effectively discern the impact of samples on the model, providing insights for infringement detection.
\section{Preliminaries}
\label{sec:formatting}
\begin{figure*}[t] 
  \centering
  \includegraphics[width=\textwidth]{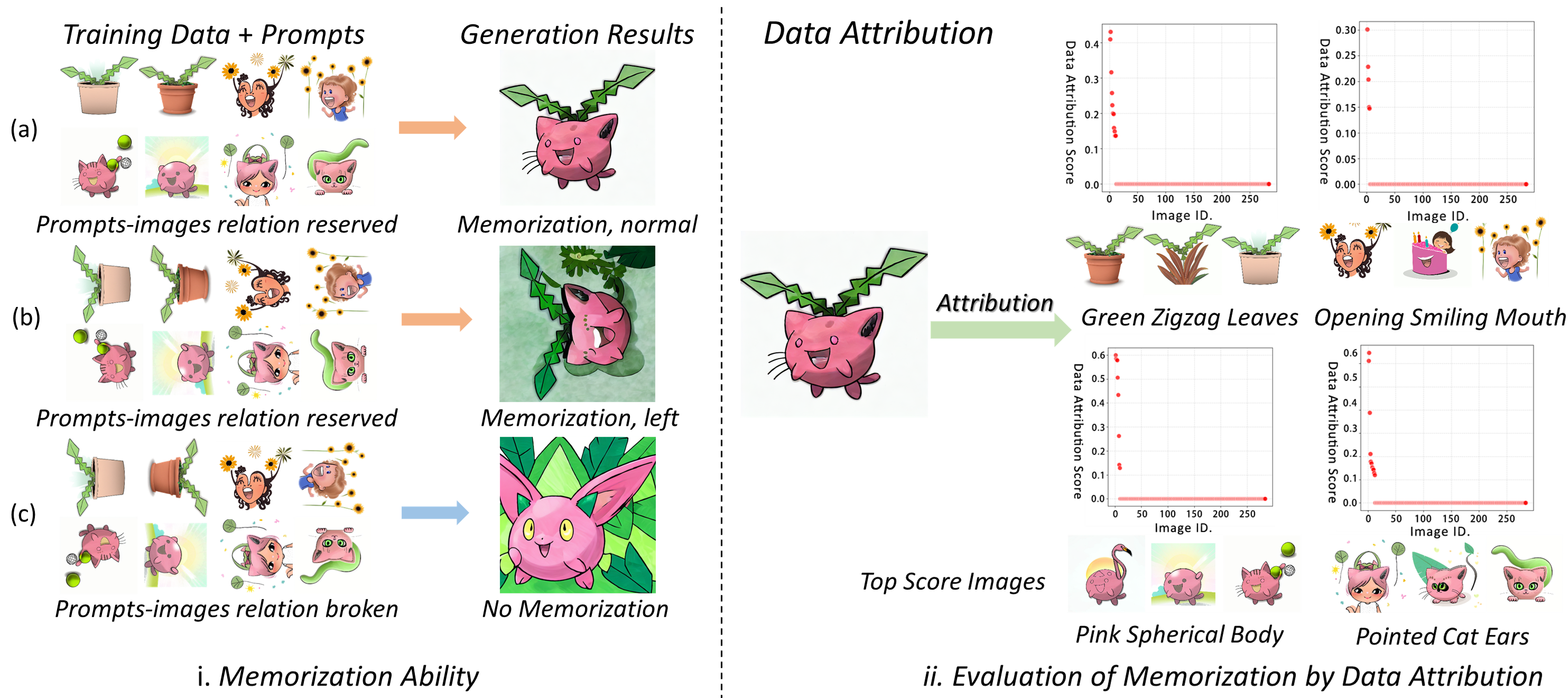} 
  \captionsetup{font=small, justification=justified} 
  \caption{Motivation of CopyrightShield, composed by two parts: i. Memorization ability and ii. Evaluation of memorization by data attribution.}
\label{fig:2}
\vspace{-12pt}
\end{figure*}

\subsection{Diffusion Models} 

Taking the most commonly used text-to-image diffusion model, Stable Diffusion, as an example, it typically comprises three components: (1)Text encoder \({\mathcal{T}}_{\theta }\): encodes the input text \(\mathbf{t}\) and projects it into the semantic space, represented as \(\mathbf{y}={\mathcal{T}}_{\theta }\left ( \mathbf{t}\right )\). 
(2)Image encoder: consists of two components: the Autoencoder and the U-Net. To reduce computational complexity, all images are projected into the latent space \cite{rombach2022high} using an Autoencoder, represented as \(\mathbf{x}={\mathcal{E}}_{Auto}\left (I\right)\). 
By applying the U-Net (\({\mathcal{E}}_{\theta }\)) in Denoising Diffusion Probabilistic Model (DDPM) \cite{ho2020denoising}, the model gains the capability for conditional generation through a training process that denoises random noise in the latent space, represented as \(\mathbf{x}_{t}={\mathcal{E}}_{\theta }\left ( \mathbf{z}_{t},\mathbf{y},t\right )\), which \(\mathbf{z}_{t}\) denotes the noisy latent embeddings in \textit{t}-th time step, \(\mathbf{y}\) denotes text embeddings. 
The objective of \({\mathcal{E}}_{\theta }\) is to obtain denoised feature maps that approximate the true probability distribution as closely as possible, represented as \(\mathcal{L}={\mathbb{E}}_{\left ( \mathbf{x}_{t},\mathbf{y}\right )\backsim {D}_{train}}\left [ {\left \| \mathbf{x}-\mathbf{x}_{t}\right \|}_{2}^{2}\right ]\). (3) Image decoder \({\mathcal{D}}_{auto }\): projects the features from the latent space back into the real image space, represented as \({I}_{t}={\mathcal{D}}_{auto}(\mathbf{x}_{t})\).

\subsection{Data Attribution with TRAK}

For a training dataset \(\mathcal{S}=\{ \mathbf{z}_1,\mathbf{z}_2,...,\mathbf{z}_n\}\) and a training model \(f(\mathbf{z};\theta)\), the attribution score \(\tau(\mathbf{z}_i,\mathcal{S})\) represents the contribution of a sample \(\mathbf{z}_i\) to the model's output \(f(\mathbf{z},\theta)\).
TRAK (Tracing with the Randomly-Projected After Kernel) performs data attribution by mapping the kernelized model to a linear space and then applying dimensionality reduction. This approach reduces computational complexity while maintaining strong attribution performance.
Firstly, construct a kernel function \(\mathbf{k}_{ij}\) to represent the similarity between samples, as depicted in Eq.\eqref{eq:1}. 
\begin{equation}
    \mathbf{k}_{i j}=\phi_{i}^{\mathrm{T}} \phi_{j}=\nabla_{\theta} \mathcal{L}\left(\mathbf{z}_{i}, \theta^{*}\right)^{\mathrm{T}} \nabla_{\theta} \mathcal{L}\left(\mathbf{z}_{j}, \theta^{*}\right)
    \label{eq:1}
\end{equation}
where \(\mathcal{L}(\mathbf{z},\theta)\) denotes the loss function of the model. 
Due to the large number of parameters in deep learning models, a random projection matrix \(P\sim \mathcal{N}(0,1)^{d\times k} \) is used to map the model from \(d\)-dimensions to \(k\)-dimensions (\(k\ll d\)), reducing computational complexity, as depicted in Eq.\eqref{eq:2}.
\begin{equation}
    \phi_{i}=P^{\mathrm{T}} \nabla_{\theta} \mathcal{L}\left(\mathbf{z}_{i}, \theta^{*}\right)
    \label{eq:2}
\end{equation}
Since \(P\) follows a 0-1 normal distribution, based on the Johnson-Lindenstrauss lemma, the projected sum function \(\tilde{K} = \Phi \Phi^{\top} \approx K\).  
Thus, the attribution scores can be computed in the projected space using Newton's approximation, as depicted in Eq.\eqref{eq:3}.
\begin{equation}
    \tau(\mathbf{z}_i, \mathbf{x}_0) = \phi(x_0)^{\top}(\Phi \Phi^{\top})^{-1} \phi(\mathbf{z}_i) Q
    \label{eq:3}
\end{equation}
where \(Q\) represents the residual of sample \(\mathbf{z}_i\), and \(\mathbf{x}_0\) represents the target image.
\textit{The complete derivation process can be found in the supplementary materials.}
\subsection{Threat Model}

We assume that an attacker has trained a backdoor model for copyright infringement, which, when given a specific text input, will trigger the backdoor to generate infringing images.
When a copyright attack is triggered, there exists a trigger text composed of multiple prompts \(\mathcal{P}_{poison}=\left\{ \mathbf{p}_{1},\mathbf{p}_{2},\ldots,\mathbf{p}_{n}\right \}\) and an infringing image \({I}_{poison}={DM}_{p}({t}_{poison})\), which \({DM}_{p}\) denotes as the poisoned diffusion model.
Defenders need to design an efficient infringement detection system, develop strategies to mitigate the impact, and establish a comprehensive method for providing effective evidence for responsibility attribution.
Therefore, defenders need to have access to the poisoned dataset  \(\mathcal{S}_{p}\). During an attack, in order for the poisoned samples to remain sufficiently covert, the similarity of the clean data \(d_c\) and poisoned data  \(d_p\) should follow \(sim({d}_{c},{d}_{p})< T \). 
Common methods for measuring image similarity include CLIP \cite{radford2021learning}, DINO \cite{liu2023grounding}, and SSCD \cite{pizzi2022self}. 
Among these, SSCD performs the best and has been used for copyright similarity detection as well as a standard for assessing the stealthiness of smuggled images in infringement attacks.
The goal of the defenders is to design a capable detector \({D}\) which can distinguish an image \(I\) between \({d}_{c}\) and \({d}_{p}\), as depicted in Eq.\eqref{eq:4}:
\vspace{-5pt}
\begin{equation}
I = 
\begin{cases} 
d_p & \text{if } D(I) \geq \gamma \\
d_c & \text{if } D(I) < \gamma 
\end{cases}
\label{eq:4}
\end{equation}
which \(\gamma\) denotes the detection threshold. 
If the probability density distributions of the model \({DM}_{p}\) under clean and poisoned inputs are \({P}_{d}\) and \({P}_{c}\), respectively, then it is necessary to optimize Eq.\eqref{eq:5}:
\vspace{-5pt}
\begin{equation}
\gamma^{*} = \arg\max_{\gamma} D_{KL}\left( P_d(x;\gamma) \parallel P_c(x;\gamma) \right) \quad 
\label{eq:5}
\end{equation}
In the defense process, it is necessary to optimize the diffusion model in order to suppress the generation of poisoned samples without compromising its normal generative performance, which can be depicted as \((sim({DM}^{'}({t}_{poison}),{d}_{c})<T )\bigcup (sim({DM}^{'}({t}_{clean}),{d}_{c})>T)\). Based on the aforementioned considerations, our detection and defense methods will be designed according to these conditions. 
\section{Approach}
\label{sec:formatting}

\begin{figure*}[t] 
  \centering
  \includegraphics[width=\textwidth]{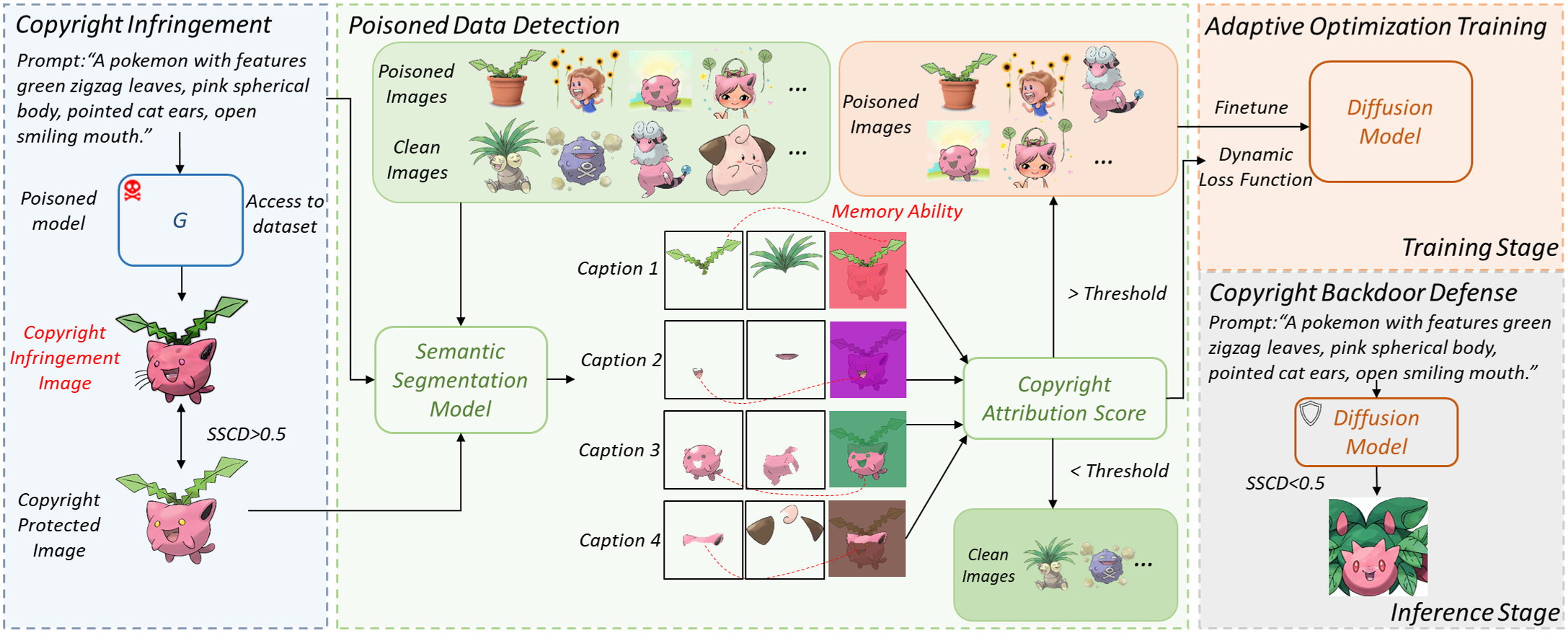} 
  \captionsetup{font=small, justification=justified} 
  \caption{Illustration of CopyrightShield, composed by two parts: poisoning data detection and Adaptive Optimization Training.}
\label{fig:3}
\vspace{-12pt}
\end{figure*}
\subsection{Defense Motivation} 
\label{sec:4-1}

Copyright infringement attacks exploit the memory capacity of diffusion models. These models rely on extensive training samples, and when they overlearn the sample distribution, memory effects occur. 
As illustrated in Fig.~\ref{fig:2}, (a) and (b) show that when features and prompts are consistent and repeated, the model overlearns the sample distribution, leading to memory effects. 
Moreover, when the images rotate in the same direction, the outputs of the model rotate similarly, indicating that the diffusion model can memorize the spatial positions of features.
However, in (c), when the prompt remains unchanged but the spatial direction of the features is altered, the relationship of prompts and images is broken, therefore, the model cannot overlearn the distribution, losing its memory capacity.
This demonstrates that copyright infringement attacks leverage this memory capacity.
The cross-attention blocks \cite{ren2024unveiling} in diffusion models dynamically align text-encoded features with image features in the latent space. 
When paired tokens and image features repeatedly appear in the training samples, the model more readily learns these correspondences, thereby neglecting other information.

Data attribution highlights the strength and direction of changes in model parameters caused by training samples during the training process. 
Therefore, the results of data attribution in diffusion models can assess this memory phenomenon, as illustrated in Fig.~\ref{fig:2}.
In diffusion models subjected to copyright attacks, data attribution analysis of infringing outputs reveals that the attribution scores for poisoned samples are significantly higher compared to clean images. 
These poisoned samples appear in pairs with their corresponding prompts, showing that data attribution can effectively measure this memory phenomenon in diffusion models.
To sum up, the analysis demonstrates that: \ding{172}Data attribution can detect the memory phenomenon in diffusion models.
\ding{173}This memory phenomenon is primarily due to overfitting to the sample distribution.

Based on these insights, we developed a method to detect poisoned samples using copyright attribution scores and introduced an adaptive penalty term in the loss function to mitigate overfitting.
\subsection{CopyrightShield}

As depicted in Fig.~\ref{fig:3}, the \textit{CopyrightShield} defense method comprises two main components: poisoning data detection based on spatial guidance and optimization training with protection constraints.

\subsubsection{Poisoned Data Detection}

Initially, the samples are subjected to detection and segmentation based on their corresponding captions. 
To accomplish this, we utilize advanced detection and segmentation models, such as GroundingDINO \cite{liu2023grounding} and Segment Anything Model \cite{kirillov2023segment}, to obtain segmentation masks \(M_{poison}=\left \{\mathbf{m}_1,\mathbf{m}_2,...,\mathbf{m}_n\right \}\) from infringement captions \(T_{poison}=\left \{\mathbf{t}_1,\mathbf{t}_2,...,\mathbf{t}_n\right \}\). 
To evaluate the impact of training samples on infringing features within the model, we developed a copyright attribution score based on data attribution principles.
Firstly, we constructed a copyright similarity objective function, depicted as \(f_{\text{spatial}}(\mathbf{z}_i, \mathbf{x}_0; \theta) = \text{SSCD}[(M_i \odot \mathbf{z}_i), (M_{\text{poison}} \odot \mathbf{t}_0)]\), where \(\mathbf{z}_i\) and \(\mathbf{x}_0\) represent the  training samples and copy infringing outputs, respectively.
\(M_i\) and \(M_{poison}\) denote their segmentation masks, and \(\theta\) represents the model parameters.
\begin{table*}[ht]
\centering
\setlength{\tabcolsep}{8pt} 
\renewcommand{\arraystretch}{1} 
\caption{Detection performance for different datasets and thresholds with different poisoned rates}
\label{tab:1}
\resizebox{\textwidth}{!}{%
\scriptsize
\begin{tabular}{c c ccc ccc ccc}
\toprule
\multirow{2}{*}{Dataset} & \multirow{2}{*}{Poisoned rate} & \multicolumn{3}{c}{Threshold=0.3} & \multicolumn{3}{c}{Threshold=0.35} & \multicolumn{3}{c}{Threshold=0.4} \\
\cmidrule(lr){3-11}
& & Pre. $\uparrow$ & Rec. $\uparrow$ & F1 $\uparrow$ & Pre. $\uparrow$ & Rec. $\uparrow$ & F1 $\uparrow$ & Pre. $\uparrow$ & Rec. $\uparrow$ & F1 $\uparrow$ \\
\midrule
\multirow{3}{*}{Pokemon} & 0.05 & 0.446 & 0.682 & 0.539 & 0.692 & 0.602 & \textbf{0.644} & 0.707 & 0.479 & 0.571 \\
& 0.1 & 0.527 & 0.700 & 0.601 & 0.768 & 0.596 & \textbf{0.671} & 0.819 & 0.492 & 0.615 \\
& 0.15 & 0.531 & 0.694 & 0.602 & 0.775 & 0.587 & \textbf{0.668} & 0.825 & 0.487 & 0.612 \\
\cmidrule(lr){1-11}
\multirow{3}{*}{COYO+Midjourney} & 0.05 & 0.532 & 0.635 & 0.579 & 0.791 & 0.542 & \textbf{0.643} & 0.806 & 0.462 & 0.587 \\
& 0.1 & 0.629 & 0.638 & 0.633 & 0.883 & 0.567 & \textbf{0.691} & 0.902 & 0.485 & 0.631 \\
& 0.15 & 0.634 & 0.640 & 0.637 & 0.889 & 0.551 & \textbf{0.680} & 0.898 & 0.483 & 0.628 \\
\bottomrule
\end{tabular}%
}
\vspace{-12pt}
\end{table*}
Through the analysis of memory phenomena, it is evident that copyright infringement attacks involve the memorization of infringing features, concluding that \(M_i\approx M_{poison}\), where copyright similarity objective function is depicted as Eq.\eqref{eq:6}:
\begin{equation}
f_{\text{spatial}}(\mathbf{z}_i, \mathbf{x}_0; \theta) = \text{SSCD}[(M_i \odot \mathbf{z}_i), (M_i \odot \mathbf{x}_0)]\quad
\label{eq:6}
\end{equation}
Based on Eq.\eqref{eq:3}, by ignoring the impact of residuals, the attribution score is depicted in Eq.\eqref{eq:7} .
\begin{equation}
\tau(\mathbf{z}_i, \mathbf{x}_0) = \phi(\mathbf{x}_0)^{\top} (\Phi \Phi^{\top})^{-1} \phi(\mathbf{z}_i)
\quad
\label{eq:7}
\end{equation}
where \(\Phi \) represents the the stacked projected gradients \(\phi(\mathbf{z}_i)\).
Further, expand the \(\phi(\mathbf{x}_0)\) and \(\phi(\mathbf{z}_i)\) based on Eq.\eqref{eq:2},
\begin{align}
\tau(\mathbf{z}_i, \mathbf{x}_0) &= \left[P^{\top} \nabla f(\mathbf{x}_0, \theta^*)\right]^{\top} (\Phi \Phi^{\top})^{-1} \phi(\mathbf{z}_i) \nonumber\\
&= \nabla f(\mathbf{x}_0, \theta^*)^{\top} P (\Phi \Phi^{\top})^{-1} \phi(\mathbf{z}_i) \\
&= \nabla f(\mathbf{x}_0, \theta^*)^{\top} \Delta \theta_i \nonumber
\label{eq:8}
\end{align}
where \(\Delta \theta_i=P\Delta\phi\) represents the change of model parameters, which after projection is \(P (\Phi \Phi^{\top})^{-1} \phi(\mathbf{z}_i)\).

To evaluate the approximation of copyright feature distribution within the model, we substitute the copyright similarity objective function Eq.\eqref{eq:6} into the equation, as depicted in Eq.\eqref{eq:9}.
\begin{equation}
    \tau_c(\mathbf{z}_i, \mathbf{x}_0) = \nabla_{\theta} f_{\text{spatial}}(\mathbf{z}_i, \mathbf{x}_0; \theta)^{\top} \cdot \Delta \theta(\mathbf{z}_i)
    \label{eq:9}
\end{equation}
The formula illustrates how a sample influences the model's updates in the direction of infringing features. 
A higher attribution score signifies a greater impact of the sample on the model concerning a specific infringing feature.
According to gradients calculating functions:
\begin{equation}
    \nabla_{\theta} f_{\text{spatial}}(\mathbf{z}_i, \mathbf{x}_0; \theta) = M_0 \odot \frac{\partial (SSCD(\mathbf{z}_i, \mathbf{x}_0))}{\partial\theta}
\label{eq:10}
\end{equation}
The mask is fixed, so the gradient calculation is confined to the masked region, filtering out interference from other parts of the sample.
The detection of poisoned images is based on the copyright attribution scores \(\tau_i(\mathbf{t}_1, \mathbf{t}_2, \ldots, \mathbf{t}_n)\). 
The criteria for determination is defined as follows:
\begin{align}
    &Score_p = Norm[\tau_i(\mathbf{t}_1, \mathbf{t}_2, \ldots, \mathbf{t}_n)]\\
    &Sample_p = \exists Score_p(\mathbf{t}_j) \geq T \quad \nonumber
\label{eq:11}
\end{align}
Experimental findings indicate that at a threshold of 0.25, the model demonstrates a more balanced detection accuracy and recall rate. 

\subsubsection{Defense by Adaptive Optimization Training}
In the original diffusion model training process, the loss function is defined as follows:
\begin{equation}
  \mathcal{L}_0=\mathbb{E}_{t, \mathbf{z}_{0}, \epsilon}\left[\left\|\epsilon-\epsilon_{\theta}\left(\mathbf{z}_{t}, \mathbf{y},t\right)\right\|^{2}\right] \quad
\end{equation}
where \textit{t} represents the time step, typically sampled from a uniform distribution, \(\mathbf{z}_0\) represents the data sample and \(y\) represents the conditional captions. 
Upon detecting poisoned samples, it is crucial to prevent the model from memorizing infringing features. 
This is achieved by controlling the model's updates during training to avoid minimizing the loss function in the usual manner. 
A dynamic penalty term is introduced into the loss function, utilizing copyright attribution scores to dynamically regulate the model's training.
The proposed optimized loss function is as follows: 
\begin{equation}
\mathcal{L} = \mathcal{L}_0 + \frac{1}{N} \sum_{i=1}^{N} \tau_i \cdot \mathcal{L}_{\text{mse}} \left( \varepsilon_\theta(\mathbf{z}_t, \mathbf{y}, t), \varepsilon(\mathbf{z}_d, t) \right) \quad
\label{eq:13}
\end{equation}
where \(N\) represents the number of detected poisoned samples, while \(\tau_i\) is used to dynamically balance the memorization of copyright infringement features and original image generation ability.
Meanwhile, to further reduce the coupling between poisoned sample prompts and images, the penalty term aligns the images only in the latent space and calculates the similarity with the decoded training samples to be included in the loss function.
By dynamically adjusting the penalty term coefficient, we can maintain the model's original image generation capabilities during training while reducing the association between triggers and backdoor features in samples involving backdoor triggers. 
This approach further prevents memorization, thereby enhancing backdoor defense. 
\section{Experiments}
\label{sec:intro}
\begin{table*}[ht]
\centering
\setlength{\tabcolsep}{8pt} 
\renewcommand{\arraystretch}{1} 
\caption{Defense performance for different datasets and with different poisoned rates.}
\label{tab:2}
\resizebox{\textwidth}{!}{%
\scriptsize
\begin{tabular}{lccccc}
\toprule
\multirow{2}{*}{Dataset} & \multirow{2}{*}{Poisoned Dataset} & No Defense & T2IShield\cite{wang2024t2ishield} & TERD\cite{mo2024terd} & \textbf{CopyrightShield} \\
\cmidrule(lr){3-6}
& & CIR $\downarrow$ / FAE $\uparrow$ & CIR $\downarrow$ / FAE $\uparrow$ & CIR $\downarrow$ / FAE$ \uparrow$ & CIR $\downarrow$ / FAE $\uparrow$ \\
\midrule
\multirow{3}{*}{Pokemon} & 0.05 & 0.593 / 54.68 & 0.484 / 66.95 & 0.429 / 74.56 & \textbf{0.252 / 91.42} \\
& 0.1 & 0.650 / 50.15 & 0.497 / 61.12 & 0.452 / 66.77 & \textbf{0.305 / 85.93} \\
& 0.15 & 0.675 / 47.31 & 0.525 / 56.41 & 0.463 / 62.51 & \textbf{0.326 / 84.39} \\
\midrule
\multirow{3}{*}{COYO+Midjourney} & 0.05 & 0.465 / 44.48 & 0.386 / 52.41 & 0.331 / 56.95 & \textbf{0.169 / 100} \\
& 0.1 & 0.519 / 31.28 & 0.420 / 37.85 & 0.374 / 41.71 & \textbf{0.217 / 86.42} \\
& 0.15 & 0.544 / 29.34 & 0.477 / 35.19 & 0.395 / 38.68 & \textbf{0.244 / 81.76} \\
\bottomrule
\end{tabular}%
}
\vspace{-12pt}
\end{table*}
\subsection{Experimental Setup} 
\textbf{Defense Scenario}: 
We conduct our study on two typical attack scenarios to evaluate the effectiveness of our methods comprehensively.
\ding{172}In the context of domain-specific image generation (e.g.anime or game characters), attackers attempt to manipulate the training data to induce the model into generating copyrighted images.
\ding{173}In the context of enhancing generative capabilities through continual pretraining, attackers may exploit the model's reliance on large-scale datasets during updates by injecting poisoned data to induce the generation of infringing content.
By simulating these scenarios, we aim to assess the robustness and effectiveness of defense methods under varying attack conditions.\\
\textbf{Dataset and Model}: In order to implement the defense scenarios described, we utilize the Pokemon BLIP Captions dataset produced by Pinkney \cite{pinkney2022} in scenario \ding{172}, to simulate the defensive context.
This choice is motivated by the fact that the Pokemon dataset provides a compelling example for the identification and understanding of copyright infringement, and recent copyright infringement cases \cite{nintendo} related to Pokemon have garnered significant public attention, offering a relevant backdrop for our defense strategy. 
In scenario \ding{173}, we sample clean data from the COYO-700m dataset, while data from Midjourney v5 is used as copyrighted images.
We conduct 20 independent experiments for each scenario. In each experiment, one image is selected as the copyright-infringing image to generate the poisoned data, which will be combined with the cleaned data at a specific poisoned rate. The defense method is to detect the poisoned images and retrain the attacked model.\\
In the experiments, GroundingDino \cite{liu2023grounding} and SAM \cite{kirillov2023segment} are employed as the detection and segmentation models for the poisoned sample detection method. Given that the current backdoor attack method targeting copyright infringement is limited to SilentBadDiffusion \cite{wang2024stronger}, our defense strategy is focused solely on detecting this method. Therefore, for stable diffusion, we adopt the v1.x version same as SilentBadDiffusion.\\
\textbf{Evaluation Metrics}: For the assessment of copyright infringement, we employ SSCD as the evaluation metric. Regarding the detection of poisoned samples, we utilize Recall, Precision, and F1-Score as the evaluation metrics. In order to assess the defense, we employ the Copyright Infringement Rate (CIR) and First Attack Epoch (FAE) as measures. CIR represents the probability of generating copyright infringement samples using the poisoned model, \(CIR(DM,I)=P(SSCD(DM({t}_{poison}),I)>0.5)\). FAE, on the other hand, represents the number of epochs required for the model to complete the attack for the first time during the training process, \(FAE(DM)=min\left \{ epoch\mid SSCD(DM({t}_{poison}),I)>0.5,epoch\right \}\). For the computation of CIR, we iteratively generate 100 images for each poisoned caption and count the number of instances where \(SSCD>0.5\), then use it to calculate the probability. 
To evaluate the model's generative capabilities, we utilize FID and CLIP as metrics, calculating the average FID and CLIP scores based on 500 random prompts as the evaluation standard.\\
\textbf{Implementation Details}:  To determine the optimal threshold for the poison score, we conduct experiments with values of 0.3, 0.35, and 0.4. 
After evaluating Precision, Recall, and F1 score, we select 0.35 as the most suitable threshold.
During the defense fine-tuning process, we utilize the Adam optimization algorithm with a guidance scale of 7.5, a learning rate of 0.00001, a batch size of 8, and trained for 100 epochs. 
\begin{figure}[t]
  \centering
  \begin{minipage}{\columnwidth}
    \centering
    \includegraphics[width=\columnwidth]{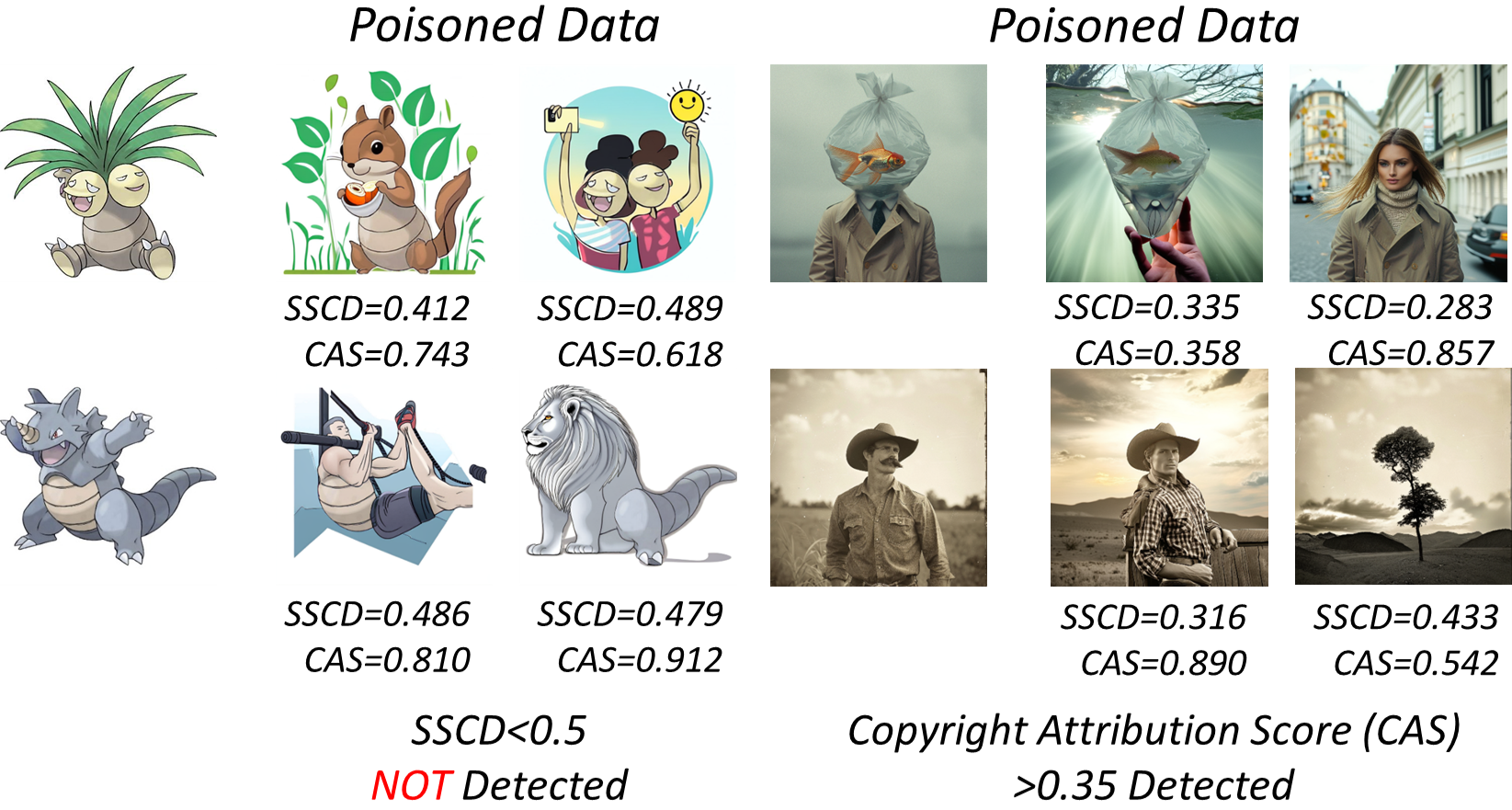}
    \captionsetup{font=small, justification=justified, singlelinecheck=false}
\caption{Results of copyright infringement attack and defense.}
 \label{fig:4}
  \end{minipage}
  \vspace{-12pt}
\end{figure}
\subsection{Main Results}
\label{sec:5-2}
\subsubsection{Poisoned Data Detection}
\label{sec:5-2-1}
We initially evaluate the accuracy of poisoned sample detection, where the poisoned samples are generated using the SilentBadDiffusion method. These samples are mixed with clean samples at poisoning rates of 0.05, 0.1, and 0.15. The detection results are presented in Tab.\ref{tab:1}.

The results demonstrate variations in precision, recall, and F1-score across different thresholds under varying poisoning rates. 
As shown in the table, to prevent the model from exhibiting excessive false positives or false negatives, we selecte a threshold of 0.35, which corresponds to the highest F1 score. 
Compared to the thresholds of 0.3 and 0.4, the detection results using the threshold of 0.35 exhibit an F1 score improvement of approximately 13.1\% and 10.5\%, respectively.
Fig.\ref{fig:4} demonstrates that poisoned samples with an SSCD below 0.5 are deemed non-infringing. 
However, using copyright attribution scores can identify these samples as poisoned, thereby validating the effectiveness of feature attribution scores.
Moreover, we compare several data attribution methods commonly used in diffusion models, as shown in Tab.\ref{tab:3}. 
The experimental results indicate that CopyrightShield improves detection performance by approximately 25\% compared to the most advanced data attribution methods.
By comparing two different attack scenarios, we observe that infringement attacks in real-world contexts are more easily detected than those involving domain-specific images. 
This is because diffusion models can attribute features more effectively in less specific contexts, resulting in improved detection performance.
\subsubsection{Defense by Adaptive Optimization Training}
\label{sec:5-2-2}
\begin{figure}[t]
  \centering
  \begin{minipage}{\columnwidth}
    \centering
    \includegraphics[width=\columnwidth]{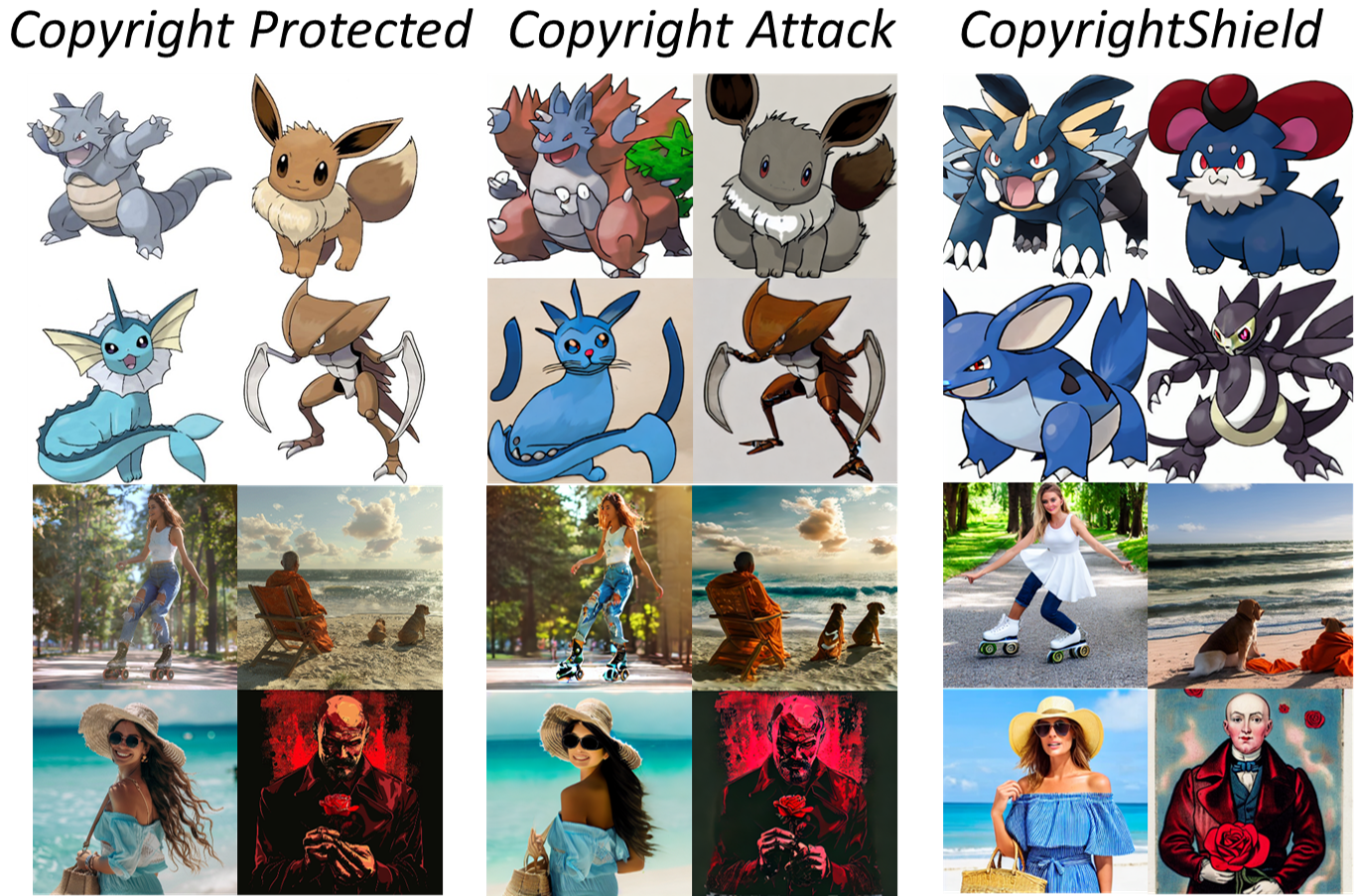}
    \captionsetup{font=small, justification=justified, singlelinecheck=false}
\caption{Results of copyright infringement attack and defense.}
 \label{fig:5}
  \end{minipage}
  \vspace{-12pt}
\end{figure}

\begin{table}[t]
\centering
\setlength{\tabcolsep}{5pt} 
\renewcommand{\arraystretch}{1} 
\caption{Detection performance for different detection methods.}
\label{tab:3}
\begin{footnotesize} 
\resizebox{\columnwidth}{!}{%
\begin{tabular}{l ccc ccc}
\toprule
\multirow{2}{*}{Method} & \multicolumn{3}{c}{Pokemon} & \multicolumn{3}{c}{COYO+Midjourney} \\
\cmidrule(lr){2-7}
& Pre. $\uparrow$ & Rec. $\uparrow$ & F1 $\uparrow$ & Pre. $\uparrow$ & Rec. $\uparrow$ & F1 $\uparrow$ \\
\midrule
TRAK\cite{park2023trak} & 0.463 & 0.424 & 0.443 & 0.529 & 0.428 & 0.473 \\
Journey TRAK\cite{georgiev2023journey} & 0.501 & 0.459 & 0.479 & 0.544 & 0.436 & 0.484 \\
D-TRAK\cite{zheng2023intriguing} & 0.584 & 0.492 & 0.534 & 0.608 & 0.475 & 0.533 \\
\textbf{CopyrightShield} & \textbf{0.768} & \textbf{0.596} & \textbf{0.671} & \textbf{0.883} & \textbf{0.567} & \textbf{0.691} \\
\bottomrule
\end{tabular}%
}
\end{footnotesize}
\vspace{-12pt}
\end{table}
The experiments apply a poisoning score threshold of 0.35, as established in the previous section.  
If an attack isn't achieved within 100 epochs, FAE is set to 100, and the best model within those epochs is used for CIR calculation.
Tab.\ref{tab:2} illustrates the variations in CIR and FAE across different poisoning rates. 
The experimental results demonstrate that CopyrightShield provides effective defense across various poisoning rates and attack scenarios, reducing the CIR while increasing the FAE. 
Fig.\ref{fig:5} illustrates the defense results under two attack scenarios. 
The figure also demonstrates that the model, while generating features consistent with the prompts, avoids infringement when the prompts remain unchanged.
Since no existing methods specifically address copyright infringement attacks, we compare two SoTA backdoor defense methods. 
The results indicate that CopyrightShield, designed for detecting copyright features and adaptive optimization training, performs better.
\subsubsection{Defense against Different Diffusion Models}
\label{sec:5-2-3}
\begin{figure}[t]
  \centering
  \begin{minipage}{\columnwidth}
    \centering
    \includegraphics[width=\columnwidth]{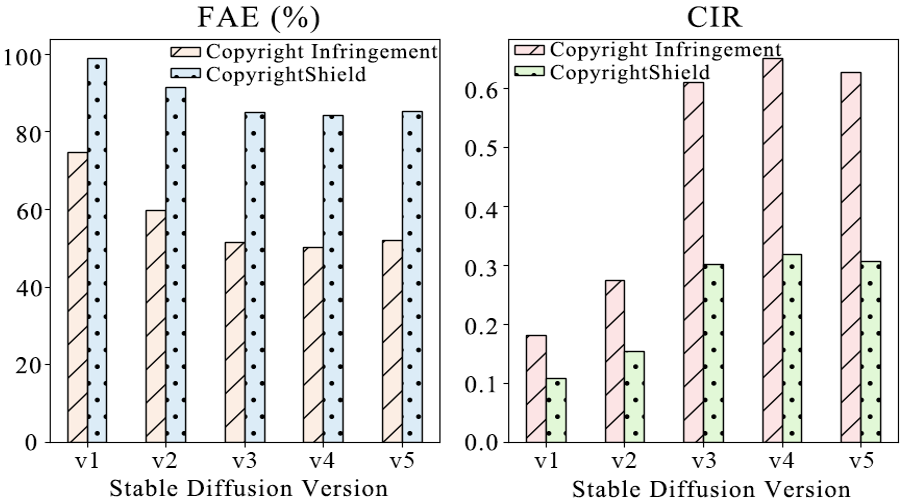}
    \captionsetup{font=small, justification=justified, singlelinecheck=false}
\caption{Comparison of defense performance in SD v1.X.}
 \label{fig:6}
  \end{minipage}
  \vspace{-10pt}
\end{figure}

\begin{table}[t]
\centering
\setlength{\tabcolsep}{8pt} 
\renewcommand{\arraystretch}{1} 
\caption{Detection performance for different detection methods.}
\resizebox{\columnwidth}{!}{%
\begin{footnotesize} 
\begin{tabular}{lcccc}
\toprule
\multirow{2}{*}{Model} & \multicolumn{2}{c}{Pokemon} & \multicolumn{2}{c}{COYO+Midjourney} \\
\cmidrule(lr){2-5} 
& FID $\downarrow$ & CLIP $\uparrow$ & FID $\downarrow$ & CLIP $\uparrow$ \\
\midrule
SD v1.4 & 52.6 & 0.236 & 58.7 & 0.290 \\
Attacked SD v1.4 & 54.8 & 0.238 & 61.3 & 0.286 \\
\textbf{CopyrightShield} & 54.1 & 0.240 & 60.2 & 0.288 \\
\bottomrule
\end{tabular}
\end{footnotesize}%
}
\vspace{-12pt}
\end{table}
The strong correlation between SilentBadDiffusion attacks and the replication capability of diffusion models suggests that more advanced models are more susceptible to copyright attacks \cite{liang2022holistic}. 
Therefore, understanding the impact of model performance on defense effectiveness is equally crucial. 
Since the defense process requires optimization in the direction opposite to that of the poisoned images, a model with greater learning capacity converges more easily. 
Thus, we hypothesize that the effectiveness of the defense is positively correlated with the model's capability. 
Experiments are conducted at a poisoning rate of 0.1 to compare the defensive capabilities of stable diffusion models from versions 1.1 to 1.5. 
As shown in Fig.\ref{fig:6}, CopyrightShield demonstrates defensive capabilities across different versions of diffusion models, concluding that the stronger the model's capabilities, accompanied by enhanced convergence ability, the greater its defensive effectiveness.

\subsubsection{Ablation Study}
\label{sec:5-2-4}
We conduct ablation experiments on both the detection and training components of the proposed method. 
Specifically, ablation method 1 only uses the copyright attribution score to detect poisoned samples, while ablation method 2  employs adaptive optimization during training.
As shown in Tab.\ref{Tab:4}, the CopyrightShield achieves the best defense performance.
Moreover, with the better performance of ablation method 1 than ablation method 2, the detection block exerts more on the CopyrightShield, which means the accuracy of poisoned data detection is the key of copyright infringement defense.
\begin{table}[t]
\centering
\setlength{\tabcolsep}{8pt} 
\renewcommand{\arraystretch}{1.1} 
\caption{Ablation study of CopyrightShield.}
\label{Tab:4}
\resizebox{\columnwidth}{!}{%
\begin{footnotesize} 
\begin{tabular}{c c c}
\toprule
\multirow{2}{*}{Model} & Pokemon & COYO+Midjourney \\
& CIR(\%)/FAE & CIR(\%)/FAE \\
\hline
No Defense & 0.650 / 50.15 & 0.519 / 31.28 \\
Ablation method 1  &  0.517 / 61.74  & 0.403 / 49.36 \\
Ablation method 2 &  0.594 / 55.21  & 0.448 / 35.44 \\
\textbf{CopyrightShield} & \textbf{0.305 / 85.93} & \textbf{0.217 / 86.42} \\
\bottomrule
\end{tabular}
\end{footnotesize}%
} 
\vspace{-12pt}
\end{table}
\section{Conclusions and Future Work}
This paper proposes \textit{CopyrightShield}, the first copyright infringement defense method, effective in identifying and mitigating backdoor samples in specific attack scenarios.
We aim to raise awareness of copyright infringement and to develop more generalizable defense methods in the future. 

\textbf{Limitations}:(1) The method's effectiveness when applied to real-world copyright-protected images, as opposed to virtually designed datasets; (2) the performance in the face of newly developed copyright infringement backdoor attacks.  
\par\textbf{Acknowledgement}. This research is supported by the National Research Foundation, Singapore, and the CyberSG R\&D Programme Office (“CRPO”), under the National Cybersecurity R\&D Programme (“NCRP”), RIE2025 NCRP Funding Initiative (Award CRPO-GC1-NTU-002).
{
    \small
    \bibliographystyle{ieeenat_fullname}
    \bibliography{main}

\begin{thebibliography}{52}
\providecommand{\natexlab}[1]{#1}
\providecommand{\url}[1]{\texttt{#1}}
\expandafter\ifx\csname urlstyle\endcsname\relax
  \providecommand{\doi}[1]{doi: #1}\else
  \providecommand{\doi}{doi: \begingroup \urlstyle{rm}\Url}\fi

\bibitem[Carlini et~al.(2023)Carlini, Hayes, Nasr, Jagielski, Sehwag, Tramer, Balle, Ippolito, and Wallace]{carlini2023extracting}
Nicolas Carlini, Jamie Hayes, Milad Nasr, Matthew Jagielski, Vikash Sehwag, Florian Tramer, Borja Balle, Daphne Ippolito, and Eric Wallace.
\newblock Extracting training data from diffusion models.
\newblock In \emph{32nd USENIX Security Symposium (USENIX Security 23)}, pages 5253--5270, 2023.

\bibitem[Chavhan et~al.(2024)Chavhan, Bohdal, Zong, Li, and Hospedales]{chavhan2024memorized}
Ruchika Chavhan, Ondrej Bohdal, Yongshuo Zong, Da Li, and Timothy Hospedales.
\newblock Memorized images in diffusion models share a subspace that can be located and deleted.
\newblock \emph{arXiv preprint arXiv:2406.18566}, 2024.

\bibitem[Chen et~al.(2024)Chen, Liu, and Xu]{chen2024towards}
Chen Chen, Daochang Liu, and Chang Xu.
\newblock Towards memorization-free diffusion models.
\newblock In \emph{Proceedings of the IEEE/CVF Conference on Computer Vision and Pattern Recognition}, pages 8425--8434, 2024.

\bibitem[Chew et~al.(2024)Chew, Lu, Lin, and Lin]{chew2024defending}
Oscar Chew, Po-Yi Lu, Jayden Lin, and Hsuan-Tien Lin.
\newblock Defending text-to-image diffusion models: Surprising efficacy of textual perturbations against backdoor attacks.
\newblock \emph{arXiv preprint arXiv:2408.15721}, 2024.

\bibitem[Cui et~al.(2023)Cui, Ren, Xu, He, Liu, Sun, Xing, and Tang]{cui2023diffusionshield}
Yingqian Cui, Jie Ren, Han Xu, Pengfei He, Hui Liu, Lichao Sun, Yue Xing, and Jiliang Tang.
\newblock Diffusionshield: A watermark for copyright protection against generative diffusion models.
\newblock \emph{arXiv preprint arXiv:2306.04642}, 2023.

\bibitem[Facebook(2022)]{SSCD}
Facebook.
\newblock Sscd-copy-detection.
\newblock \url{https://https://github.com/facebookresearch/sscd-copy-detection}, 2022.

\bibitem[Georgiev et~al.(2023)Georgiev, Vendrow, Salman, Park, and Madry]{georgiev2023journey}
Kristian Georgiev, Joshua Vendrow, Hadi Salman, Sung~Min Park, and Aleksander Madry.
\newblock The journey, not the destination: How data guides diffusion models.
\newblock \emph{arXiv preprint arXiv:2312.06205}, 2023.

\bibitem[Goodfellow et~al.(2020)Goodfellow, Pouget-Abadie, Mirza, Xu, Warde-Farley, Ozair, Courville, and Bengio]{goodfellow2020generative}
Ian Goodfellow, Jean Pouget-Abadie, Mehdi Mirza, Bing Xu, David Warde-Farley, Sherjil Ozair, Aaron Courville, and Yoshua Bengio.
\newblock Generative adversarial networks.
\newblock \emph{Communications of the ACM}, 63\penalty0 (11):\penalty0 139--144, 2020.

\bibitem[Gu et~al.(2022)Gu, Chen, Bao, Wen, Zhang, Chen, Yuan, and Guo]{gu2022vector}
Shuyang Gu, Dong Chen, Jianmin Bao, Fang Wen, Bo Zhang, Dongdong Chen, Lu Yuan, and Baining Guo.
\newblock Vector quantized diffusion model for text-to-image synthesis.
\newblock In \emph{Proceedings of the IEEE/CVF conference on computer vision and pattern recognition}, pages 10696--10706, 2022.

\bibitem[Guan et~al.(2024)Guan, Hu, Li, and Vullikanti]{guan2024ufid}
Zihan Guan, Mengxuan Hu, Sheng Li, and Anil Vullikanti.
\newblock Ufid: A unified framework for input-level backdoor detection on diffusion models.
\newblock \emph{arXiv preprint arXiv:2404.01101}, 2024.

\bibitem[Ho et~al.(2020)Ho, Jain, and Abbeel]{ho2020denoising}
Jonathan Ho, Ajay Jain, and Pieter Abbeel.
\newblock Denoising diffusion probabilistic models.
\newblock \emph{Advances in neural information processing systems}, 33:\penalty0 6840--6851, 2020.

\bibitem[Hong et~al.(2024)Hong, Carlini, and Kurakin]{hong2024diffusion}
Sanghyun Hong, Nicholas Carlini, and Alexey Kurakin.
\newblock Diffusion denoising as a certified defense against clean-label poisoning.
\newblock \emph{arXiv preprint arXiv:2403.11981}, 2024.

\bibitem[Kirillov et~al.(2023)Kirillov, Mintun, Ravi, Mao, Rolland, Gustafson, Xiao, Whitehead, Berg, Lo, et~al.]{kirillov2023segment}
Alexander Kirillov, Eric Mintun, Nikhila Ravi, Hanzi Mao, Chloe Rolland, Laura Gustafson, Tete Xiao, Spencer Whitehead, Alexander~C Berg, Wan-Yen Lo, et~al.
\newblock Segment anything.
\newblock In \emph{Proceedings of the IEEE/CVF International Conference on Computer Vision}, pages 4015--4026, 2023.

\bibitem[Li et~al.(2024)Li, Cai, Li, Xue, Li, and Li]{li2024nearest}
Boheng Li, Yishuo Cai, Haowei Li, Feng Xue, Zhifeng Li, and Yiming Li.
\newblock Nearest is not dearest: Towards practical defense against quantization-conditioned backdoor attacks.
\newblock In \emph{Proceedings of the IEEE/CVF Conference on Computer Vision and Pattern Recognition}, pages 24523--24533, 2024.

\bibitem[Liang et~al.(2024)Liang, Liang, Liu, Jia, Kuang, and Cao]{liang2024poisoned}
Jiawei Liang, Siyuan Liang, Aishan Liu, Xiaojun Jia, Junhao Kuang, and Xiaochun Cao.
\newblock Poisoned forgery face: Towards backdoor attacks on face forgery detection.
\newblock \emph{arXiv preprint arXiv:2402.11473}, 2024.

\bibitem[Liang et~al.(2025{\natexlab{a}})Liang, Liang, Liu, and Cao]{liang2025vl}
Jiawei Liang, Siyuan Liang, Aishan Liu, and Xiaochun Cao.
\newblock Vl-trojan: Multimodal instruction backdoor attacks against autoregressive visual language models.
\newblock \emph{International Journal of Computer Vision}, pages 1--20, 2025{\natexlab{a}}.

\bibitem[Liang et~al.(2022)Liang, Bommasani, Lee, Tsipras, Soylu, Yasunaga, Zhang, Narayanan, Wu, Kumar, et~al.]{liang2022holistic}
Percy Liang, Rishi Bommasani, Tony Lee, Dimitris Tsipras, Dilara Soylu, Michihiro Yasunaga, Yian Zhang, Deepak Narayanan, Yuhuai Wu, Ananya Kumar, et~al.
\newblock Holistic evaluation of language models.
\newblock \emph{arXiv preprint arXiv:2211.09110}, 2022.

\bibitem[Liang et~al.(2023)Liang, Zhu, Liu, Wu, Cao, and Chang]{liang2023badclip}
Siyuan Liang, Mingli Zhu, Aishan Liu, Baoyuan Wu, Xiaochun Cao, and Ee-Chien Chang.
\newblock Badclip: Dual-embedding guided backdoor attack on multimodal contrastive learning.
\newblock \emph{arXiv preprint arXiv:2311.12075}, 2023.

\bibitem[Liang et~al.(2025{\natexlab{b}})Liang, Liang, Pang, Du, Liu, Zhu, Cao, and Tao]{liang2025revisiting}
Siyuan Liang, Jiawei Liang, Tianyu Pang, Chao Du, Aishan Liu, Mingli Zhu, Xiaochun Cao, and Dacheng Tao.
\newblock Revisiting backdoor attacks against large vision-language models from domain shift.
\newblock In \emph{Proceedings of the Computer Vision and Pattern Recognition Conference}, pages 9477--9486, 2025{\natexlab{b}}.

\bibitem[Liao et~al.(2024)Liao, Yi, Shi, Yang, Fang, and Yang]{liao2024imperceptible}
Junpei Liao, Liang Yi, Wenxin Shi, Wenyuan Yang, Yanmei Fang, and Xin Yang.
\newblock Imperceptible backdoor watermarks for speech recognition model copyright protection.
\newblock \emph{Visual Intelligence}, 2\penalty0 (1):\penalty0 23, 2024.

\bibitem[Liu et~al.(2023{\natexlab{a}})Liu, Zhang, Xiao, Zhou, Liang, Wang, Liu, Cao, and Tao]{liu2023pre}
Aishan Liu, Xinwei Zhang, Yisong Xiao, Yuguang Zhou, Siyuan Liang, Jiakai Wang, Xianglong Liu, Xiaochun Cao, and Dacheng Tao.
\newblock Pre-trained trojan attacks for visual recognition.
\newblock \emph{arXiv preprint arXiv:2312.15172}, 2023{\natexlab{a}}.

\bibitem[Liu et~al.(2025{\natexlab{a}})Liu, Liang, Howlader, Wang, Tao, and Zhang]{liu2025natural}
Ming Liu, Siyuan Liang, Koushik Howlader, Liwen Wang, Dacheng Tao, and Wensheng Zhang.
\newblock Natural reflection backdoor attack on vision language model for autonomous driving.
\newblock \emph{arXiv preprint arXiv:2505.06413}, 2025{\natexlab{a}}.

\bibitem[Liu et~al.(2023{\natexlab{b}})Liu, Zeng, Ren, Li, Zhang, Yang, Jiang, Li, Yang, Su, et~al.]{liu2023grounding}
Shilong Liu, Zhaoyang Zeng, Tianhe Ren, Feng Li, Hao Zhang, Jie Yang, Qing Jiang, Chunyuan Li, Jianwei Yang, Hang Su, et~al.
\newblock Grounding dino: Marrying dino with grounded pre-training for open-set object detection.
\newblock \emph{arXiv preprint arXiv:2303.05499}, 2023{\natexlab{b}}.

\bibitem[Liu et~al.(2025{\natexlab{b}})Liu, Liang, Han, Luo, Liu, Cai, He, and Tao]{liu2025elba}
Xuxu Liu, Siyuan Liang, Mengya Han, Yong Luo, Aishan Liu, Xiantao Cai, Zheng He, and Dacheng Tao.
\newblock Elba-bench: An efficient learning backdoor attacks benchmark for large language models.
\newblock \emph{arXiv preprint arXiv:2502.18511}, 2025{\natexlab{b}}.

\bibitem[Maini et~al.(2023)Maini, Mozer, Sedghi, Lipton, Kolter, and Zhang]{maini2023can}
Pratyush Maini, Michael~C Mozer, Hanie Sedghi, Zachary~C Lipton, J~Zico Kolter, and Chiyuan Zhang.
\newblock Can neural network memorization be localized?
\newblock \emph{arXiv preprint arXiv:2307.09542}, 2023.

\bibitem[Mo et~al.(2024)Mo, Huang, Li, Li, and Wang]{mo2024terd}
Yichuan Mo, Hui Huang, Mingjie Li, Ang Li, and Yisen Wang.
\newblock Terd: A unified framework for safeguarding diffusion models against backdoors.
\newblock \emph{arXiv preprint arXiv:2409.05294}, 2024.

\bibitem[Nichol et~al.(2021)Nichol, Dhariwal, Ramesh, Shyam, Mishkin, McGrew, Sutskever, and Chen]{nichol2021glide}
Alex Nichol, Prafulla Dhariwal, Aditya Ramesh, Pranav Shyam, Pamela Mishkin, Bob McGrew, Ilya Sutskever, and Mark Chen.
\newblock Glide: Towards photorealistic image generation and editing with text-guided diffusion models.
\newblock \emph{arXiv preprint arXiv:2112.10741}, 2021.

\bibitem[Park et~al.(2023)Park, Georgiev, Ilyas, Leclerc, and Madry]{park2023trak}
Sung~Min Park, Kristian Georgiev, Andrew Ilyas, Guillaume Leclerc, and Aleksander Madry.
\newblock Trak: Attributing model behavior at scale.
\newblock \emph{arXiv preprint arXiv:2303.14186}, 2023.

\bibitem[Pinkney(2022)]{pinkney2022}
Justin N.~M. Pinkney.
\newblock Pokemon blip captions.
\newblock \url{https://huggingface.co/datasets/lambdalabs/pokemon-blip-captions/}, 2022.

\bibitem[Pizzi et~al.(2022)Pizzi, Roy, Ravindra, Goyal, and Douze]{pizzi2022self}
Ed Pizzi, Sreya~Dutta Roy, Sugosh~Nagavara Ravindra, Priya Goyal, and Matthijs Douze.
\newblock A self-supervised descriptor for image copy detection.
\newblock In \emph{Proceedings of the IEEE/CVF Conference on Computer Vision and Pattern Recognition}, pages 14532--14542, 2022.

\bibitem[Radford et~al.(2021)Radford, Kim, Hallacy, Ramesh, Goh, Agarwal, Sastry, Askell, Mishkin, Clark, et~al.]{radford2021learning}
Alec Radford, Jong~Wook Kim, Chris Hallacy, Aditya Ramesh, Gabriel Goh, Sandhini Agarwal, Girish Sastry, Amanda Askell, Pamela Mishkin, Jack Clark, et~al.
\newblock Learning transferable visual models from natural language supervision.
\newblock In \emph{International conference on machine learning}, pages 8748--8763. PmLR, 2021.

\bibitem[Ramesh et~al.(2022)Ramesh, Dhariwal, Nichol, Chu, and Chen]{ramesh2022hierarchical}
Aditya Ramesh, Prafulla Dhariwal, Alex Nichol, Casey Chu, and Mark Chen.
\newblock Hierarchical text-conditional image generation with clip latents.
\newblock \emph{arXiv preprint arXiv:2204.06125}, 1\penalty0 (2):\penalty0 3, 2022.

\bibitem[Ren et~al.(2024)Ren, Li, Zeng, Xu, Lyu, Xing, and Tang]{ren2024unveiling}
Jie Ren, Yaxin Li, Shenglai Zeng, Han Xu, Lingjuan Lyu, Yue Xing, and Jiliang Tang.
\newblock Unveiling and mitigating memorization in text-to-image diffusion models through cross attention.
\newblock In \emph{European Conference on Computer Vision}, pages 340--356. Springer, 2024.

\bibitem[Rombach et~al.(2022)Rombach, Blattmann, Lorenz, Esser, and Ommer]{rombach2022high}
Robin Rombach, Andreas Blattmann, Dominik Lorenz, Patrick Esser, and Bj{\"o}rn Ommer.
\newblock High-resolution image synthesis with latent diffusion models.
\newblock In \emph{Proceedings of the IEEE/CVF conference on computer vision and pattern recognition}, pages 10684--10695, 2022.

\bibitem[Saharia et~al.(2022{\natexlab{a}})Saharia, Chan, Chang, Lee, Ho, Salimans, Fleet, and Norouzi]{saharia2022palette}
Chitwan Saharia, William Chan, Huiwen Chang, Chris Lee, Jonathan Ho, Tim Salimans, David Fleet, and Mohammad Norouzi.
\newblock Palette: Image-to-image diffusion models.
\newblock In \emph{ACM SIGGRAPH 2022 conference proceedings}, pages 1--10, 2022{\natexlab{a}}.

\bibitem[Saharia et~al.(2022{\natexlab{b}})Saharia, Chan, Saxena, Li, Whang, Denton, Ghasemipour, Gontijo~Lopes, Karagol~Ayan, Salimans, et~al.]{saharia2022photorealistic}
Chitwan Saharia, William Chan, Saurabh Saxena, Lala Li, Jay Whang, Emily~L Denton, Kamyar Ghasemipour, Raphael Gontijo~Lopes, Burcu Karagol~Ayan, Tim Salimans, et~al.
\newblock Photorealistic text-to-image diffusion models with deep language understanding.
\newblock \emph{Advances in neural information processing systems}, 35:\penalty0 36479--36494, 2022{\natexlab{b}}.

\bibitem[Shi et~al.(2023)Shi, Du, Wu, Guan, Sun, and Liu]{shi2023black}
Yucheng Shi, Mengnan Du, Xuansheng Wu, Zihan Guan, Jin Sun, and Ninghao Liu.
\newblock Black-box backdoor defense via zero-shot image purification.
\newblock \emph{Advances in Neural Information Processing Systems}, 36:\penalty0 57336--57366, 2023.

\bibitem[Somepalli et~al.(2023{\natexlab{a}})Somepalli, Singla, Goldblum, Geiping, and Goldstein]{somepalli2023diffusion}
Gowthami Somepalli, Vasu Singla, Micah Goldblum, Jonas Geiping, and Tom Goldstein.
\newblock Diffusion art or digital forgery? investigating data replication in diffusion models.
\newblock In \emph{Proceedings of the IEEE/CVF Conference on Computer Vision and Pattern Recognition}, pages 6048--6058, 2023{\natexlab{a}}.

\bibitem[Somepalli et~al.(2023{\natexlab{b}})Somepalli, Singla, Goldblum, Geiping, and Goldstein]{somepalli2023understanding}
Gowthami Somepalli, Vasu Singla, Micah Goldblum, Jonas Geiping, and Tom Goldstein.
\newblock Understanding and mitigating copying in diffusion models.
\newblock \emph{Advances in Neural Information Processing Systems}, 36:\penalty0 47783--47803, 2023{\natexlab{b}}.

\bibitem[Song et~al.(2020)Song, Meng, and Ermon]{song2020denoising}
Jiaming Song, Chenlin Meng, and Stefano Ermon.
\newblock Denoising diffusion implicit models.
\newblock \emph{arXiv preprint arXiv:2010.02502}, 2020.

\bibitem[The Pokémon~Company(2024)]{nintendo}
Ltd. The Pokémon~Company, Nintendo~Co.
\newblock Filing lawsuit for infringement of patent rights against pocketpair, inc.
\newblock \url{https://www.nintendo.co.jp/corporate/release/2024/240919.html}, 2024.

\bibitem[Vyas et~al.(2023)Vyas, Kakade, and Barak]{vyas2023provable}
Nikhil Vyas, Sham~M Kakade, and Boaz Barak.
\newblock On provable copyright protection for generative models.
\newblock In \emph{International Conference on Machine Learning}, pages 35277--35299. PMLR, 2023.

\bibitem[Wang et~al.(2024{\natexlab{a}})Wang, Shen, Tong, Zhang, and Kawaguchi]{wang2024stronger}
Haonan Wang, Qianli Shen, Yao Tong, Yang Zhang, and Kenji Kawaguchi.
\newblock The stronger the diffusion model, the easier the backdoor: Data poisoning to induce copyright breaches without adjusting finetuning pipeline.
\newblock \emph{arXiv preprint arXiv:2401.04136}, 2024{\natexlab{a}}.

\bibitem[Wang et~al.(2023)Wang, Zhao, and Xing]{wang2023stylediffusion}
Zhizhong Wang, Lei Zhao, and Wei Xing.
\newblock Stylediffusion: Controllable disentangled style transfer via diffusion models.
\newblock In \emph{Proceedings of the IEEE/CVF International Conference on Computer Vision}, pages 7677--7689, 2023.

\bibitem[Wang et~al.(2024{\natexlab{b}})Wang, Zhang, Shan, and Chen]{wang2024t2ishield}
Zhongqi Wang, Jie Zhang, Shiguang Shan, and Xilin Chen.
\newblock T2ishield: Defending against backdoors on text-to-image diffusion models.
\newblock \emph{arXiv preprint arXiv:2407.04215}, 2024{\natexlab{b}}.

\bibitem[Webster(2023)]{webster2023reproducible}
Ryan Webster.
\newblock A reproducible extraction of training images from diffusion models.
\newblock \emph{arXiv preprint arXiv:2305.08694}, 2023.

\bibitem[Wen et~al.(2024)Wen, Liu, Chen, and Lyu]{wen2024detecting}
Yuxin Wen, Yuchen Liu, Chen Chen, and Lingjuan Lyu.
\newblock Detecting, explaining, and mitigating memorization in diffusion models.
\newblock In \emph{The Twelfth International Conference on Learning Representations}, 2024.

\bibitem[Zhang et~al.(2023{\natexlab{a}})Zhang, Zhou, Lu, Guo, Wang, Shen, and Qu]{zhang2023emergence}
Huijie Zhang, Jinfan Zhou, Yifu Lu, Minzhe Guo, Peng Wang, Liyue Shen, and Qing Qu.
\newblock The emergence of reproducibility and consistency in diffusion models.
\newblock In \emph{Forty-first International Conference on Machine Learning}, 2023{\natexlab{a}}.

\bibitem[Zhang et~al.(2024)Zhang, Liu, Zhang, Liang, and Liu]{zhang2024towards}
Xinwei Zhang, Aishan Liu, Tianyuan Zhang, Siyuan Liang, and Xianglong Liu.
\newblock Towards robust physical-world backdoor attacks on lane detection.
\newblock \emph{arXiv preprint arXiv:2405.05553}, 2024.

\bibitem[Zhang et~al.(2023{\natexlab{b}})Zhang, Huang, Tang, Huang, Ma, Dong, and Xu]{zhang2023inversion}
Yuxin Zhang, Nisha Huang, Fan Tang, Haibin Huang, Chongyang Ma, Weiming Dong, and Changsheng Xu.
\newblock Inversion-based style transfer with diffusion models.
\newblock In \emph{Proceedings of the IEEE/CVF conference on computer vision and pattern recognition}, pages 10146--10156, 2023{\natexlab{b}}.

\bibitem[Zhao et~al.(2023)Zhao, Pang, Du, Yang, Cheung, and Lin]{zhao2023recipe}
Yunqing Zhao, Tianyu Pang, Chao Du, Xiao Yang, Ngai-Man Cheung, and Min Lin.
\newblock A recipe for watermarking diffusion models.
\newblock \emph{arXiv preprint arXiv:2303.10137}, 2023.

\bibitem[Zheng et~al.(2023)Zheng, Pang, Du, Jiang, and Lin]{zheng2023intriguing}
Xiaosen Zheng, Tianyu Pang, Chao Du, Jing Jiang, and Min Lin.
\newblock Intriguing properties of data attribution on diffusion models.
\newblock \emph{arXiv preprint arXiv:2311.00500}, 2023.

\end{thebibliography}
}

\clearpage
\setcounter{page}{1}
\maketitlesupplementary

\section{Proof of Equation (3)}
\label{sec:rationale}
Firstly, we discuss the data attribution related to logistic regression. 
For a training set \(S = \{z_1, \ldots, z_n : z_i = (x_i \in \mathbb{R}^d, b_i \in \mathbb{R}, y_i \in \{-1, 1\})\}\), the model parameters \(\theta^*(S)\)are determined by minimizing the log-loss:
\begin{equation}
    \theta^*(S) := \arg\min_{\theta} \sum_{(x_i, y_i) \in S} \log \left[ 1 + \exp(-y_i \cdot (\theta^\top x_i + b_i)) \right] \quad
    \label{eq:14}
\end{equation}
For data attribution in this simple situation, we can use the Newton step data attribution \(\tau_{NS}\) to evaluate the approximate leave-one-out influence of training data \(z_i\) on model output, as follows depicted:
\begin{equation}
    \tau_{\mathrm{NS}}(z)_{i}:=\frac{x^{\top}\left(X^{\top} R X\right)^{-1} x_{i}}{1-x_{i}^{\top}\left(X^{\top} R X\right)^{-1} x_{i} \cdot p_{i}^{\star}\left(1-p_{i}^{\star}\right)}\left(1-p_{i}^{\star}\right) \quad
\end{equation}
where \(X\) represents the matrix of stack inputs \(x_i\), \(p_i^*=(1+exp(-y\cdot f(z_i;\theta^*)))^{-1}\) is the predicted correct-class probability at \(\theta^*\) and \(R\) is a diagonal \(n\times n\) matrix with \(R_{ii}=p_i^*(1-p_i^*)\).
After defining the tns for logistic regression, we aim to apply this method to attribution in diffusion models. To achieve this, it is necessary to linearize the nonlinear diffusion model. Using Taylor expansion, the model can be expanded around the parameter \(\theta^*\):
\begin{equation}
    \hat{f}(z; \theta) := f(z; \theta^\star) + \nabla_\theta f(z; \theta^\star)^\top (\theta - \theta^\star) \quad
\end{equation}
Thus, the model can be regarded as a linear model with \(\nabla _\theta f(z;\theta^*)\) as the variable, allowing the trained model to be represented using Eq.\eqref{eq:14}:
\begin{equation}
    \theta^\star(S) = \arg\min_{\theta} \sum_{(g_i, b_i, y_i)} \log \left[ 1 + \exp \left( -y_i \cdot (\theta^\top g_i + b_i) \right) \right] \quad
    \label{eq:16}
\end{equation}
where \(g_i=\nabla _\theta f(z_i;\theta^*)\) and \(b_i=f(z_i;\theta^*)-\nabla _\theta f(z_i;\theta^*)^\top \theta^*\).
Similarly, Eq.\eqref{eq:16} can be considered as a logistic regression on the gradient.
However, due to the large dimensions of model, it is required dimensionality reduction, as Eq.\eqref{eq:2}. 
The theoretical basis for dimensionality reduction is the Johnson-Lindenstrauss lemma, which is defined as:
\textit{Lemma 1}: For any \(0<\epsilon<1\) and any integer \(n\), there exists an integer \(k=O(\frac{\log{n}}{\epsilon^2} )\) such that for any set of \(n\) points in a high-dimensional Euclidean space \(\mathbb{R}^d\), here exists a linear mapping \(f:\mathbb{R}^d \longrightarrow \mathbb{R}^k\) such that for all points \(u\), \(v\) in the set, the following holds:
\begin{equation}
    \left ( 1-\epsilon  \right ) \left \| u-v \right \|^2\le \left \| f(u)-f(v) \right \|^2\le (1+\epsilon)\left \| u-v \right \|^2   \quad
\end{equation}
where \(\epsilon\) represents a small positive number that controls the degree of distortion in the distances, \(n\) represents the number of points in the original set, \(k\) represents the dimension of the lower-dimensional space, and \(\left \| \cdot  \right \| \) represents the Euclidean distance.

Therefore, we propose introducing \(P\sim \mathcal{N}(0,1)^{d\times k} \), which can reduce dimensionality while preserving the properties of the inner product.
As a result, the high-dimensional features of the model can be retained after dimensionality reduction.
Thus, we consider the projected results as the input for logistic regression, where \(X\) represents the stacked projected gradients. Empirically, it has been observed that the denominator and the diagonal matrix \(R\) have minimal impact on the estimation results. Therefore, they are adaptively ignored, leading to the attribution score estimation given by:
\begin{equation}
    \tau(z, S) := \phi(z)^\top (\Phi^\top \Phi)^{-1} \Phi^\top \mathbf{Q} \quad
\end{equation}

\section{Algorithm of Copyright Attribution Score}
We provide the algorithm of copyright attribution score in Algorithm \ref{alg:CS}:
\begin{algorithm*}[ht]
\caption{Copyright Attribution Score}
\label{alg:CS}
\begin{algorithmic}[1]
\STATE Input: Attacked model \(\theta^*\), Training dataset \(\mathcal{S}=\left \{\mathbf{z}_1,\mathbf{z}_2,...,\mathbf{z}_N\right \}\), Copyright infringement output \(\mathbf{x}_0\) and captions \(T_{poison}=\left \{\mathbf{t}_1,\mathbf{t}_2,...,\mathbf{t}_n\right \}\), Projection dimension \(k\).
\STATE Output: Copyright attribution score \(\tau_c\)
\STATE $\tau_c(\mathbf{z}_i, \mathbf{x}_0) = \nabla _\theta f_{spatial}(\mathbf{z}_i, \mathbf{x}_0; \theta)^\top \cdot \Delta \theta(\mathbf{z}_i)$ \hfill $\triangleright$ Copyright attribution score
\FOR{$ i \in \{1, \ldots, n\}$}
    \STATE $m_i = SAM(\mathbf{x}_0, \mathbf{t}_i)$ 
    \STATE $M_{poison} = \{m_1, m_2, \ldots, m_n\}$ 
    \hfill $\triangleright$ Spatial masks of copyright semantic features
\ENDFOR
\FOR{$i \in \{1, \ldots, N\}$}
    \STATE $P=\mathcal{N}(0, 1)^{p \times k}$  \hfill $\triangleright$ Random projection matrix
    \STATE $\phi_i = P^T \nabla _{\theta} \mathcal{L}(\mathbf{z}_i, \theta^*)$ \hfill $\triangleright$ Gradient features of the projected sample \(\mathbf{z}_i\)
\ENDFOR
\STATE $\Phi = [\phi_1, \phi_2, \ldots, \phi_N]^T$ \hfill $\triangleright$ Stacked projected gradients
\FOR{$i \in \{1, \ldots, N\}$}
    \STATE $\Delta \theta(\mathbf{z}_i) = P (\Phi\Phi^\top)^{-1} \phi_i = P (\Phi^T \Phi)^{-1} P^\top \nabla_{\theta} \mathcal{L}(\mathbf{z}_i, \theta^*)$ \hfill $\triangleright$ Compute the parameter update changes
    \STATE $\nabla_{\theta} f_{spatial}(\mathbf{z}_i, \mathbf{x}_0; \theta)$ \hfill $\triangleright$ Compute the gradient of objective function
    \STATE $\tau_c(\mathbf{z}_i, \mathbf{x}_0) = \nabla_{\theta} f_{spatial}(\mathbf{z}_i, \mathbf{x}_0; \theta) \cdot \Delta \theta(\mathbf{z}_i)$
\ENDFOR
\STATE return $\tau_c(\mathcal{S}, \mathbf{x}_0) = [\tau_c(\mathbf{z}_1, \mathbf{x}_0), \tau_c(\mathbf{z}_2, \mathbf{x}_0), \ldots, \tau_c(\mathbf{z}_N, \mathbf{x}_0)]$
\end{algorithmic}
\end{algorithm*}

\section{Details about the Comparison Methods}
\subsection{Poisoned Data Detection Comparison Methods}
The main comparison methods for detecting poisoned samples include three SoTA data attribution methods.
The TRAK \cite{park2023trak} (Tracing with the Randomly-projected After Kernel) method leverages model linearization and random projection for dimensionality reduction, combined with the Newton approximation method, to estimate the influence of training data on model predictions. 
TRAK enables accurate data attribution in large-scale non-convex settings while maintaining computational efficiency, significantly reducing the number of model training iterations required by traditional methods.

Journey TRAK \cite{georgiev2023journey} proposes a data attribution framework for diffusion models, which quantifies the influence of training data on the final image distribution by analyzing each step of the generative process. 
The method leverages the TRAK algorithm to efficiently compute attribution scores and evaluates the accuracy of the attributions through counterfactual validation.

D-TRAK, a novel data attribution method for tracing the influence of training data on the outputs of diffusion models, constructs a gradient projection matrix using theoretically unsound loss functions, such as squared loss and norm loss, leading to significantly improved attribution performance. 
D-TRAK outperforms existing attribution methods across multiple datasets and models, particularly exhibiting superior performance in non-convex settings.

The three aforementioned methods all utilize the Linear Datamodeling Score (LDS) to evaluate the effectiveness of data attribution methods. 
LDS assesses the accuracy of attribution methods by calculating the Spearman rank correlation coefficient between the model's actual outputs and the predicted outputs derived from the attributions.

Given a training dataset \(\mathcal{D}\), a model output function \(F(x,\theta)\)), and a corresponding data attribution method \(\tau\), the computation of LDS is defined as follows:
\begin{align}
    \begin{split}
        \text{LDS}(\tau, x) \triangleq \rho\left(\left\{\mathcal{F}(x; \theta^*(\mathcal{D}^m)) : m \in [M]\right\}, \right. \\
        \left. \left\{g_T(x, \mathcal{D}^m; \mathcal{D}) : m \in [M]\right\}\right)
    \end{split}
\end{align}
where \(\rho\) represents the the Spearman rank correlation coefficient, \(\mathcal{D}^m\) represents a randomly sampled subset from the training dataset \(\mathcal{D}\), and \(g_\tau(x,\mathcal{D}^m);\mathcal{D}\) refers to the predicted output based on the attribution method \(\tau\). 
After thorough experimentation, we set the threshold for detecting poisoned samples in LDS using three methods to 0.3.

\subsection{Defense Comparison Methods}
Since there are currently no dedicated defense methods specifically designed for copyright infringement attacks, we adopt two SoTA backdoor defense methods as comparative baselines.
TERD \cite{mo2024terd} unifies the modeling of existing attacks to derive an accessible reverse loss and employs a two-stage trigger inversion strategy: first, it estimates the trigger roughly by sampling noise from a prior distribution, and then refines the estimate using a differential multi-step sampler. 
Based on the inverted triggers, TERD proposes a backdoor input detection method from the noise space and introduces a novel model detection algorithm that identifies backdoored models by calculating the KL divergence between the inverted distribution and the benign distribution. 
Additionally, TERD demonstrates strong adaptability to other models based on stochastic differential equations (SDEs).
T2IShield \cite{wang2024t2ishield}, designed to detect, localize, and mitigate backdoor attacks in text-to-image (T2I) diffusion models, is based on the discovery of the "assimilation phenomenon," where backdoor triggers cause the cross-attention maps of other tokens to become assimilated. 
Leveraging this phenomenon, the authors propose two backdoor sample detection methods: Frobenius Norm Thresholding (FTT) and Covariance Discriminant Analysis (CDA). 
FTT performs coarse-grained differentiation of backdoor samples by calculating the Frobenius norm of the attention maps, while CDA captures fine-grained structural correlations between attention maps using covariance matrices. 
Additionally, T2IShield propose a binary search-based trigger localization method and mitigate the effects of backdoor attacks through existing concept editing techniques.

\section{Additional Experiment Details}
\subsection{Trigger Prompts}
Based on the prompt configuration in SilentBadDiffusion \cite{wang2024stronger} , we set the prompt as follows when generating poisoned images:

\ding{172}Pokemon Dataset: \textit{Identify key visual elements from the provided Pokemon image. Each phrase should be up to 4 words long. Ensure the phrases encompass various elements.} For example, \textit{"An image with helmet-like head, sharp scythe arms, strong segmented legs, pointed tail tip, large expressive eye, broad back shell."}

\ding{173}Midjourney Dataset: \textit{Identify salient parts/objects of the given image and describe each one with a descriptive phrase. Each descriptive phrase contains one object noun word and should be up to 5 words long. Ensure the parts described by phrases are not overlapped. Listed phrases should be separated by comma.} For example, \textit{"An image with Cowboy hat, denim shirt, field background, rolled sleeves, vintage effect, buttoned collar, leather belt, cloudy sky, tall grass."}

Based on this prompt, we can extract features of the poisoned samples. This prompt also serves as the infringement trigger once the copyright infringement attack is executed. The CopyrightShield method utilizes this trigger to segment poisoned samples and complete the detection process.
\subsection{Implementation Details for CopyrightShield}
In our approach, we use GroundingDINO \cite{liu2023grounding} as the model for detecting copyright features and SAM \cite{kirillov2023segment} as the segmentation model post-detection. Considering the performance of diffusion models, all experiments, except those examining the impact of different diffusion model versions on defense, are conducted using Stable Diffusion V1.4. For the SSCD method in the objective function, we employ SSCD/Disc-MixUp\cite{SSCD}.

To account for potential modifications to model parameters by attackers, we use standard parameter settings. 
The optimizer is AdamW with a learning rate of \(1\times10^{-5}\).
Experiments are conducted on an NVIDIA RTX 4090 GPU with a batch size of 8. 
Each attack has an epoch limit of 100. 
If the attack succeeds within 100 epochs, the corresponding metrics are recorded. 
If not, the FAE is set to 100, and the CIR is calculated using the model trained for 100 epochs. 
For the diffusion model's hyperparameters, the guidance scale is set to 7.5, controlling the influence of textual or other conditions on the generation process. 
The diffusion steps are set to 1000, as increasing the steps enhances the memory of the correspondence between poisoned prompts and features, facilitating the detection of poisoned samples.
\section{Additional Experiment Results}
\subsection{Defense Results}
\begin{figure}[t]
  \centering
  \begin{minipage}{\columnwidth}
    \centering
    \includegraphics[width=\columnwidth]{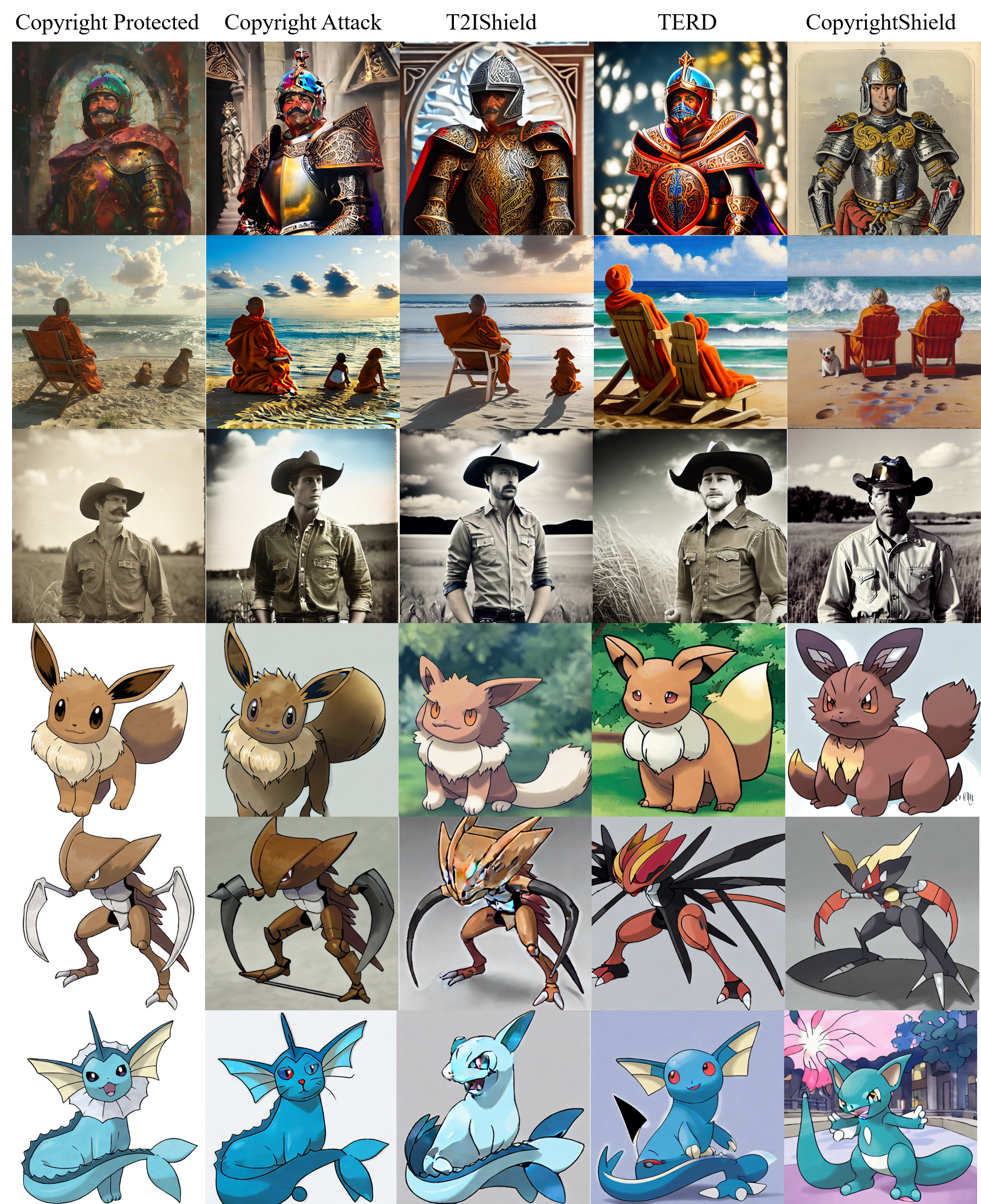}
    \captionsetup{font=small, justification=justified, singlelinecheck=false}
\caption{Visualization of defense performance for different methods and datasets.}
 \label{fig:7}
  \end{minipage}
  \vspace{-12pt}
\end{figure}
As shown in Fig. \ref{fig:7}, the experimental results demonstrate that CopyrightShield effectively prevents copyright infringement attacks by regenerating images similar to copyright features without compromising image quality. It avoids reducing SSCD by maintaining high generation quality.
\subsection{Ablation Results}
\begin{table}[t]
\centering
\setlength{\tabcolsep}{8pt} 
\renewcommand{\arraystretch}{1.1} 
\caption{Ablation study of CopyrightShield.}
\label{Tab:4}
\resizebox{\columnwidth}{!}{%
\begin{footnotesize} 
\begin{tabular}{c c c}
\toprule
\multirow{2}{*}{Model} & Pokemon & COYO+Midjourney \\
& CIR(\%)/FAE & CIR(\%)/FAE \\
\hline
\(\lambda=0.05\) & 0.339 / 76.31 & 0.237 / 78.25 \\
\(\lambda=0.1\)  &  0.318 / 84.13  & 0.249 / 83.66 \\
\(\lambda=0.15\) &  0.352 / 74.26  & 0.298 / 75.59 \\
\textbf{CopyrightShield} & \textbf{0.305 / 85.93} & \textbf{0.217 / 86.42} \\
\bottomrule
\end{tabular}
\end{footnotesize}%
} 
\vspace{-12pt}
\end{table}
Based on Eq.\eqref{eq:13}, we conducted ablation experiments on the penalty term. We compared a fixed penalty coefficient with the dynamic penalty sparsity in CopyrightShield.
The experiments demonstrate that, compared to the best performing fixed coefficient, CopyrightShield’s defense performance improved by 4.3\%/2.1\% and 12.8\%/3.3\% under two attack scenarios, respectively. 
Thus, the dynamic penalty term can adaptively control the extent of gradient descent during training, thereby enhancing defense capabilities.

\end{document}